%% file: LEGNet.tex
\documentclass[10pt,twocolumn,letterpaper]{article}

\usepackage{iccv}              

\usepackage{colortbl}
\usepackage{booktabs} 
\usepackage{graphicx}
\usepackage{multirow}
\usepackage{makecell}
\usepackage{pifont}
\usepackage{caption}
\usepackage{wrapfig}
\usepackage{amsmath,amsfonts}
\usepackage{color}
\usepackage{xcolor}


\definecolor{iccvblue}{rgb}{0.21,0.49,0.74}
\usepackage[pagebackref,breaklinks,colorlinks,allcolors=iccvblue]{hyperref}

\def\paperID{ 5 } 
\def\confName{ICCV Workshop}
\def\confYear{2025}

\title{LEGNet: A Lightweight Edge-Gaussian Network for Low-Quality Remote Sensing Image Object Detection}

\author{
	Wei Lu\textsuperscript{1},
	Si-Bao Chen\textsuperscript{1}\thanks{Si-Bao Chen is corresponding author.},
	Hui-Dong Li\textsuperscript{1},
	Qing-Ling Shu\textsuperscript{1},
	Chris H. Q. Ding\textsuperscript{2},
	Jin Tang\textsuperscript{1},
	Bin Luo\textsuperscript{1} \\
	\textsuperscript{1}Anhui University, 
	\textsuperscript{2}The Chinese University of Hong Kong (Shenzhen) \\
	\tt\small 
	\{2858191255, 2563489133\}@qq.com, \{sbchen, tangjin, luobin\}@ahu.edu.cn,\\
	\tt\small 
	e23301341@stu.ahu.edu.cn, chrisding@cuhk.edu.cn
}

\begin{document}
	\maketitle
	\input{sec/0_abstract}    
	\input{sec/1_intro}

	\input{sec/2_rework}
	\input{sec/3methods}

\input{sec/4exp}

	\input{sec/5con} 
	\textbf{Acknowledgements:}
	This work was supported in part by NSFC Key Project of International (Regional) Cooperation and Exchanges (No. 61860206004), NSFC Key Project of Joint Fund for Enterprise Innovation and Development (No. U20B2068, U24A20342) and National Natural Science Foundation of China (No. 61976004).

	{
		\small
		\bibliographystyle{ieeenat_fullname}
		\bibliography{legnet}
	}
	
	
	\clearpage 
	\onecolumn  
	
	\begin{center}
		\Large\bf
		Supplementary Materials for ``LEGNet: A Lightweight Edge-Gaussian Network for Low-Quality Remote Sensing Image Object Detection''
	\end{center}
	\vspace{1.5em} 
	
	\setcounter{page}{1} 
	\appendix  
	
	This supplementary document provides additional details to complement the main paper. It aims to facilitate reproducibility and deepen the understanding of our proposed LEGNet, including its architectural design, core components, and extensive experimental validation.

	The appendix is organized as follows:
	\begin{itemize}
		\item \textbf{Appendix A} provides detailed descriptions of the datasets used for evaluation and the experimental setup.
		\item \textbf{Appendix B} presents the specific architectural configurations for the different variants of LEGNet (Tiny and Small).
		\item \textbf{Appendix C} includes an in-depth discussion on our key design choices, particularly the rationale for employing explicit priors and the sensitivity of hyperparameters.
		\item \textbf{Appendix D} offers a visual analysis of intermediate feature responses, providing insight into how LEGNet enhances feature representations compared to other methods.
		\item \textbf{Appendix E} showcases additional qualitative results, visualizing detection performance on the DOTA-v1.0 benchmark.
	\end{itemize}

	\section{Datasets and Experimental Setup}
	As shown in~\cref{tab:datasets}, LEGNet was evaluated on five well-established datasets for object detection in remote sensing images: DOTA 1.0~\cite{xia2018dota}, DOTA 1.5~\cite{xia2018dota}, DIOR-R~\cite{AOPG}, FAIR1M-v1.0~\cite{sun2022fair1m}, and VisDrone2019~\cite{du2019visdronedet}. These datasets provide diverse challenges, including variations in object scale, orientation, density, and environmental conditions, making them ideal for assessing the robustness of object detection models.

	\begin{table}[h] \centering \scriptsize	
		\renewcommand{\arraystretch}{2}
		\setlength{\tabcolsep}{2mm}{
			\begin{tabular}{|l|cc|c|c|c|c|}\specialrule{0.8pt}{0pt}{0pt}
				\rowcolor[rgb]{0.92,0.92,0.92}	\textbf{Dataset} & \textbf{Train Set}& \textbf{Test Set}  & \textbf{Instances} 
				& \textbf{Categories} & \textbf{Resolution} & \textbf{Key Features} \\ \hline \hline
				DOTA-v1.0 & 21,046 &10,833& 188,282 & 15 & 1,024 $\times$ 1,024 & High-resolution, diverse object sizes, aspect ratios, orientations \\ \hline
				DOTA-v1.5& 21,046 &10,833& 403,318 & 16 &1,024 $\times$ 1,024 & Adds small object annotations and container crane category, orientations \\ \hline
				DIOR-R & 11,725 &11,738& 192,472 & 20 & 800 $\times$ 800 & high inter-class similarity, intra-class diversity, orientations \\ \hline
				FAIR1M-v1.0 & 95,396& 48,701  & $>$1,000,000 & 5 (37 sub) & \makecell[c]{682 $\times$ 682, \\ 1,024 $\times$ 1,024, \\ 2,048 $\times$ 2,048} & Large-scale, fine-grained categories, geographic metadata, orientations \\ \hline
				VisDrone2019 & 6,471&548 & $>$2,600,000 & 10 & \makecell[c]{480 $\times$ 360 \\ to \\ 2,000 $\times$ 1,500 \\ (Varies)}
				& Drone-based, dense targets, complex backgrounds, diverse scenarios \\ \specialrule{0.8pt}{0pt}{0pt}
		\end{tabular}}
		\caption{Comparison of Remote Sensing Object Detection Datasets}
		\label{tab:datasets}
	\end{table}
	
	• \textbf{DOTA-v1.0 and v1.5.} The datasets are widely recognized benchmark for object detection in aerial images, featuring high-resolution images ranging from 800 $\times$ 800 to 20,000 $\times$ 20,000 pixels. The datasets were split into training (1,411 images), validation (458 images), and test sets (937 images). To meet the DOTA-v1.0 and v1.5 benchmark, images were divided into 1,024 $\times$ 1,024 patches with a 200-pixel overlap, resulting in approximately 21,046 patches for training and 10,833 for testing. Models were trained on the combined training and validation sets, and their performance was evaluated on the test set.
	
	DOTA-v1.0 comprises 2,806 images with 188,282 annotated instances across 15 categories, including Plane (PL), Baseball diamond (BD), Bridge (BR), Ground track field (GTF), Small vehicle (SV), Large vehicle (LV), Ship (SH), Tennis court (TC), Basketball court (BC), Storage tank (ST), Soccer-ball field (SBF), Roundabout (RA), Harbor (HA), Swimming pool (SP), and Helicopter (HC). Its diversity in object sizes, orientations, and aspect ratios poses significant challenges for oriented object detection. DOTA-v1.5 builds upon DOTA-v1.0 by incorporating annotations for extremely small objects (less than 10 pixels) and introducing a new category, container crane (CC), increasing the total to 403,318 instances. This makes DOTA-v1.5 particularly suited for evaluating models on small and densely packed objects. 
	
	• \textbf{DIOR-R.} Derived from the DIOR dataset~\cite{li2020object}, DIOR-R is tailored for object detection in optical remote sensing images with oriented bounding box (OBB) annotations for precise localization. It includes 23,463 images, each at 800 $\times$ 800 pixels, with 192,472 annotated instances across 20 categories, such as airplanes, ships, and baseball fields. The dataset's high inter-class similarity and intra-class diversity challenge models to distinguish between visually similar objects.
	
	• \textbf{FAIR1M-v1.0.} FAIR1M-v1.0 is one of the largest datasets for fine-grained object detection in high-resolution remote sensing images, containing 15,266 images with over 1 million annotated instances. These are organized into 5 main categories and 37 sub-categories, with images captured at resolutions of 0.3--0.8 meters. To ensure fairness, we follow the same dataset processing approach as LSKNet~\cite{li2023large}. We adopt multi-scale training and testing strategy by first rescaling the images into three scales (0.5, 1.0, 1.5), and then cropping each scaled image into 1,024 $\times$ 1,024 sub-images with a patch overlap of 500 pixels, resulting in approximately 95,396 patches for training and 48,701 for testing. The dataset includes geographic metadata, such as latitude, longitude, and resolution, enhancing its utility for geospatial applications. Its scale and fine-grained categorization make it ideal for testing models on complex, large-scale detection tasks.

	• \textbf{VisDrone2019.} The dataset is a comprehensive benchmark for drone-based object detection, consisting of 10,209 static images captured by various drone-mounted cameras across diverse urban and rural environments. It features over 2.6 million bounding box annotations across 10 categories, including pedestrians, cars, and bicycles, under varying weather and lighting conditions. The dataset's dense target distributions and complex backgrounds make it a challenging testbed for developing robust detection algorithms for unmanned aerial vehicle (UAV) images.

	\section{LEGNet Configuration Details}
	\begin{table*}[h] \centering \scriptsize
		\setlength{\tabcolsep}{4pt}
		\renewcommand{\arraystretch}{1.3}
		\begin{tabular}{c|c|c|cc} \specialrule{0.8pt}{0pt}{0pt}
			\multirow{2}*{\textbf{Stage}} 	&\textbf{Downs.} 	&\textbf{Layer }
			&\multicolumn{2}{c}{\textbf{(Input / Output) channels (\(C\))}}\\ \cline{4-5}
			&\textbf{Rate}&\textbf{Specification}&\textbf{Tiny} &\textbf{Small}  \\ \hline
			\multirow{3}*{1}&\multirow{3}*{$\frac{H}{4}\times \frac{W}{4}$}        
			& \texttt{LoG-Stem Layer}  &$3/32$ &$3/64$  \\ \cline{3-5}
			
			&&\multirow{2}*{\texttt{[LEG Block] $\times N_1$}}  
			& \multirow{2}*{$32/32$}	& \multirow{2}*{$64/64$}   \\ 	&&&&\\	 \hline
			\multirow{3}*{2}&\multirow{3}*{$\frac{H}{8}\times \frac{W}{8}$}        
			&\texttt{DRFD Module} 	&$32/64$ & $64/128$  \\ \cline{3-5}
			
			&&\multirow{2}*{\texttt{[LEG Block] $\times N_2$}} 
			& \multirow{2}*{$64/64$}	& \multirow{2}*{$128/128$}  \\ 	&&&&\\	 \hline
			\multirow{3}*{3}&\multirow{3}*{$\frac{H}{16}\times \frac{W}{16}$}      
			&\texttt{DRFD Module} &$64/128$ &$128/256$ \\ \cline{3-5}
			
			&&\multirow{2}*{\texttt{[LEG Block] $\times N_3$}}  
			& \multirow{2}*{$128/128$}	& \multirow{2}*{$256/256$}   \\  	&&&&\\  \hline
			\multirow{3}*{4}&\multirow{3}*{$\frac{H}{32}\times \frac{W}{32}$}      
			&\texttt{DRFD Module}  &$128/256$ &$256/512$ \\ \cline{3-5}
			
			&&\multirow{2}*{\texttt{[LEG Block] $\times N_4$}}   
			& \multirow{2}*{$256/256$}	& \multirow{2}*{$512/512$}  \\ 	&&&&\\ \hline
			\multicolumn{3}{c}{Number of Block}	\vline
			&\multicolumn{2}{c}{[$N_1$, $N_2$, $N_3$, $N_4$] = [1, 4, 4, 2]} \\  \hline
		\end{tabular}
		\caption{Architecture configurations of LEGNet.} \label{tab_achitecture}	\vspace{-6pt}	
	\end{table*}
	
	\cref{tab_achitecture} provides the architectural configurations of LEGNet, which features two distinct scales (Tiny and Small) structured into four sequential stages. Each stage progressively downsamples the spatial resolution of the input feature maps.
	Each stage commences with an initial layer responsible for downsampling, followed by a series of repeated LEG Blocks ($N_i\times$).\\
	\textbf{Stage 1}.\\
	Downsampling Rate: The spatial resolution is reduced to $H/4 \times W/4$ relative to the original input dimensions.\\
	Initial Layer: A `LoG-Stem Layer' processes the input. For the Tiny configuration, it transforms 3 input channels to 32 output channels ($3/32$). For the Small configuration, it maps 3 input channels to 64 output channels ($3/64$). This indicates an initial feature extraction and channel expansion.\\
	Core Blocks: This stage includes $N_1$ repetitions of the `[LEG Block]'. Both Tiny and Small versions maintain the channel dimensions within these blocks (Tiny: $32/32$; Small: $64/64$), indicating a focus on learning hierarchical features without further channel changes at this sub-stage. \\
	\textbf{Stage 2}.\\
	Downsampling Rate: The spatial resolution is further reduced to $H/8 \times W/8$.\\
	Initial Layer: A `DRFD Module' is employed for inter-stage transition and downsampling. The Tiny model expands channels from 32 to 64 ($32/64$), while the Small model expands from 64 to 128 ($64/128$). This module incorporates downsampling operations to achieve the specified resolution reduction.\\
	Core Blocks: This stage comprises $N_2$ `[LEG Block]' repetitions. Channels are maintained within these blocks (Tiny: $64/64$; Small: $128/128$), suggesting the primary role of these blocks is to refine features at the current resolution.\\
	\textbf{Stage 3}.\\
	Downsampling Rate: The spatial resolution is reduced to $H/16 \times W/16$.\\
	Initial Layer: Another `DRFD Module' facilitates the transition. The Tiny model maps channels from 64 to 128 ($64/128$), and the Small model maps from 128 to 256 ($128/256$). This continues the pattern of channel expansion and downsampling.\\
	Core Blocks: $N_3$ `[LEG Block]' repetitions are used, preserving channel dimensions (Tiny: $128/128$; Small: $256/256$).\\
	\textbf{Stage 4}.\\
	Downsampling Rate: The final downsampling brings the resolution to $H/32 \times W/32$.\\
	Initial Layer: The last `DRFD Module' handles the transition. The Tiny model transforms channels from 128 to 256 ($128/256$), and the Small model from 256 to 512 ($256/512$).\\
	Core Blocks: $N_4$ `[LEG Block]' repetitions are present, maintaining channel dimensions (Tiny: $256/256$; Small: $512/512$).
	
	\textbf{Block Repetitions}.
	The number of `LEG Block' repetitions for each stage, denoted as $[N_1, N_2, N_3, N_4]$, is consistently set to $[1, 4, 4, 2]$ for both Tiny and Small configurations.

	\section{Discussion on Design Choices}
	This section elaborates on key design choices within LEGNet, providing the rationale for embedding explicit priors over purely end-to-end learning and discussing the sensitivity of crucial hyperparameters.
	
	\subsection{Rationale for Employing Explicit Priors}
	A natural question regarding our approach is why we chose to explicitly encode priors, such as edge and Gaussian features, rather than relying on a sufficiently deep or complex network (e.g., a Transformer-based model) to learn them automatically. While modern deep networks possess immense learning capabilities, our approach of embedding explicit priors offers several distinct advantages, particularly in the context of our goal to build a lightweight and robust backbone for RSOD.
	
	\begin{itemize}
		\item \textbf{Data and Parameter Efficiency:} Learning fundamental concepts like edges or Gaussian-like attention from scratch is a parameter-intensive task. By embedding these priors through parameter-free operators (our LoG-Stem and the fixed Scharr/Gaussian filters in the EGA module), we provide the network with a strong, built-in "head start." This makes the model significantly more data-efficient and allows it to achieve high performance with a much lower parameter count—a core objective of our lightweight LEGNet design.
		
		\item \textbf{Robustness to Degradation:} This is a cornerstone of our motivation. End-to-end models learn features based on statistical patterns in the training data. When input images are degraded (e.g., due to blur, low contrast, or noise), these patterns can become weak or distorted, causing learned filters to fail. In contrast, our explicit edge detectors are deterministic operators that can reliably extract structural information even from degraded signals. By providing the network with this robust edge map, we ensure that subsequent layers receive meaningful structural cues, enhancing the model's resilience to poor imaging conditions.
		
		\item \textbf{Guided Learning and Regularization:} Explicit priors act as a strong inductive bias, guiding the network to focus on structurally relevant information from the earliest stages. This serves as a form of regularization, discouraging the model from overfitting to spurious textures or background noise. As discussed in our macro design (\cref{lim_future}), this biases the model towards learning "the right features for the right reasons," leading to better generalization.
		
		\item \textbf{Interpretability:} The use of well-defined operators like Laplacian of Gaussian provides a degree of interpretability to the early-stage feature extraction process. We know precisely that the initial layers are enhancing edges, which aligns with human visual processing and provides clearer insight into the model's behavior, as supported by our feature visualizations in \cref{fig:intermediate_features} of the Appendix.
	\end{itemize}
	
	In summary, our choice is not based on the premise that deep networks *cannot* learn these features, but rather that explicitly encoding them provides a more efficient, robust, and targeted pathway to building a high-performing lightweight model for the specific challenges of RSOD.
	
	\subsection{Discussion on Hyperparameter Sensitivity}
	The hyperparameters of the Gaussian kernel, namely its size and bandwidth (standard deviation, $\sigma$), are influential components of the EGA module. While a full ablation study was beyond the scope of our primary investigation, we discuss their roles and the reasoning behind our choices here.
	
	\begin{itemize}
		\item \textbf{Gaussian Kernel Size:} The kernel size determines the spatial extent of the feature aggregation. A larger kernel incorporates context from a wider neighborhood, beneficial for larger objects, but risks over-smoothing details and blurring together small, dense objects. Conversely, a smaller kernel preserves fine details but may fail to capture sufficient context. Our selected kernel size was determined during preliminary experiments to strike a balance suitable for the multi-scale nature of objects in datasets like DOTA and FAIR1M. The chosen value proved effective at capturing salient features without significant information loss.
		
		\item \textbf{Gaussian Bandwidth ($\sigma$):} The bandwidth controls the decay rate of the Gaussian function. A small $\sigma$ creates a sharp filter that heavily prioritizes the central feature, while a large $\sigma$ creates a smoother filter that gives more uniform weight to the neighborhood. The value of $\sigma$ is often set proportionally to the kernel size to ensure the distribution fits naturally within the kernel's window. Our implementation follows this standard practice, a well-established heuristic in computer vision for ensuring stability.
		
	\end{itemize}
	Crucially, while these parameters are important, we found the model's performance to be robust to minor variations around our chosen values. This robustness is largely because the subsequent learnable layers (e.g., the 1x1 convolutions within our LEG Blocks) grant the network the flexibility to adapt and recalibrate the features generated by the fixed-parameter Gaussian module. Therefore, the overall architecture is not hyper-sensitive to these specific values, and the selected parameters represent a stable and effective configuration for the task.

	
	\section{Analysis of Intermediate Feature Responses} 
\begin{figure*}[h!]
		\begin{center}
			\includegraphics[width=1\linewidth]{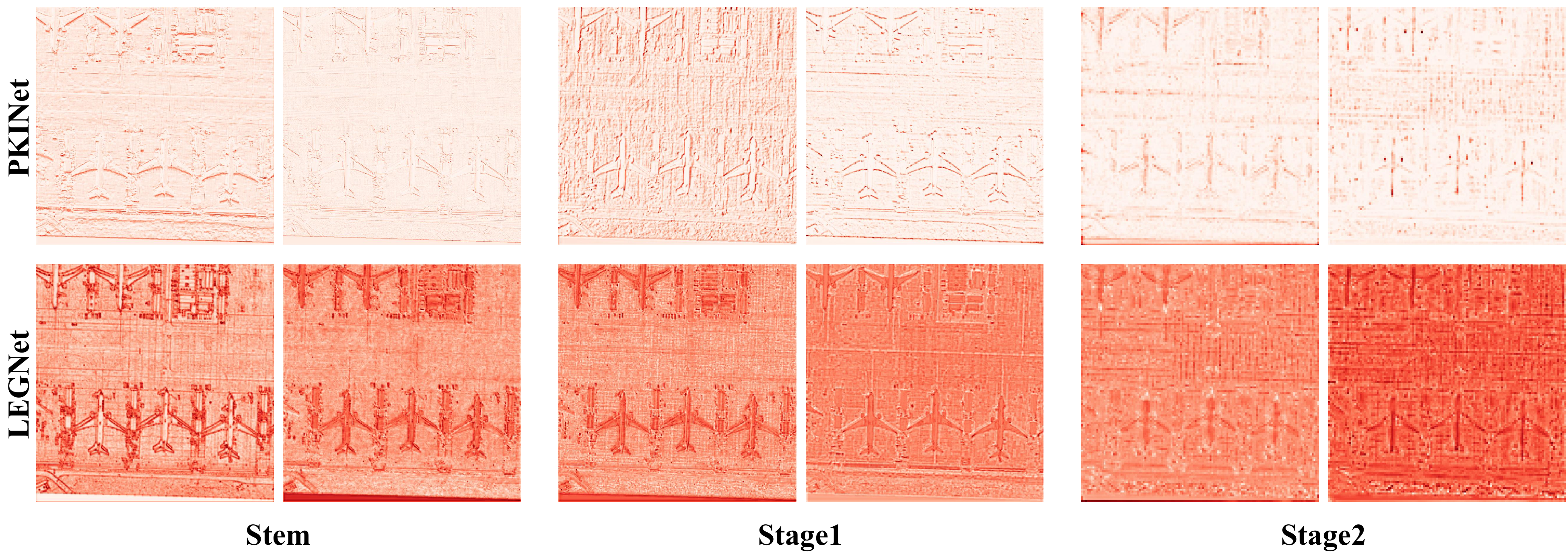}
		\end{center}
		\caption{Visualization of intermediate feature maps from PKINet-S (top row) and our LEGNet-S (bottom row). LEGNet's early stages capture more complete edge details, while Stage 2 demonstrates enhanced focus on salient object regions. This supports the effectiveness of our proposed design in refining feature representations for robust detection.}
		\label{fig:intermediate_features}
	\end{figure*}

	To provide deeper insight into LEGNet's mechanism, we visualize the intermediate feature maps. As shown in \cref{fig:intermediate_features}, we compare the feature responses of our LEGNet-S with a strong competitor, PKINet-S, at different stages of the backbone: the initial stem layer, the output of Stage 1, and the output of Stage 2. 
	
	The visualizations reveal distinct differences in feature representation. In the stem layer and Stage 1, LEGNet (bottom row) demonstrates a superior ability to extract comprehensive and complete edge information across the entire image compared to PKINet (top row). This is attributed to our LoG-Stem layer and the EGA module, which explicitly enhance structural details from the outset. As the network deepens into Stage 2, LEGNet exhibits a more focused attention on salient object features, effectively suppressing background noise and highlighting regions of interest. In contrast, the feature responses from PKINet appear more diffuse. This comparison visually substantiates our claim that by integrating explicit edge and Gaussian priors, LEGNet learns more robust and meaningful feature representations, particularly in the crucial early stages of processing. Due to the highly abstract nature of features in deeper stages (3 and 4), we focus on these initial stages where the impact of our design is most visually interpretable.
	
	\section{Visualization Results} 
	
\begin{figure*}[hb]
		\begin{center}
			\includegraphics[width=1\linewidth]{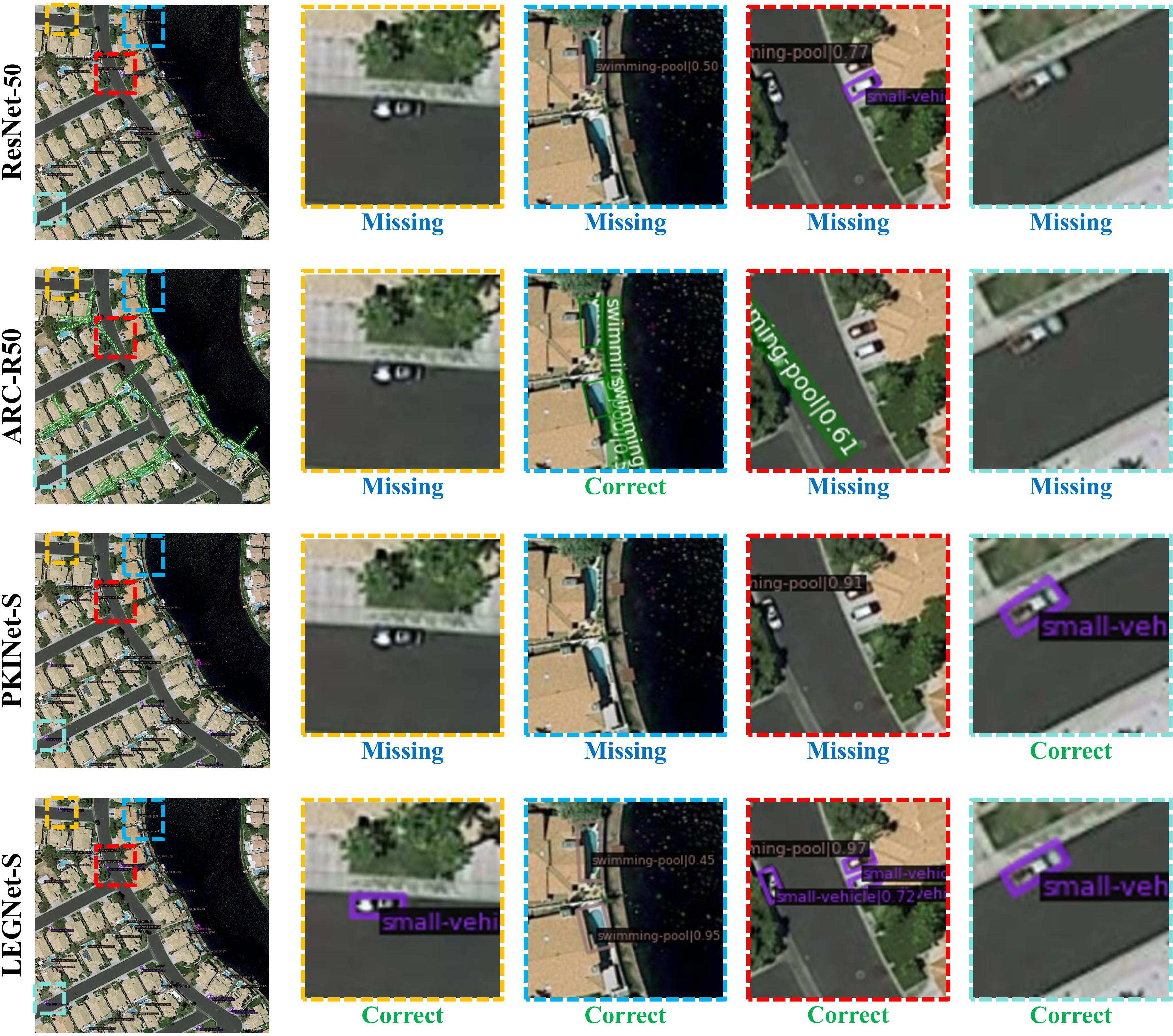}
		\end{center}
		\caption{Visualization of detection results on DOTA-v1.0 test set. Input images resolution were 1,024 $\times$ 1,024.}	\vspace{-2mm}
		\label{vis4}
	\end{figure*}

\begin{figure*}[hb]
		\begin{center}
			\includegraphics[width=1\linewidth]{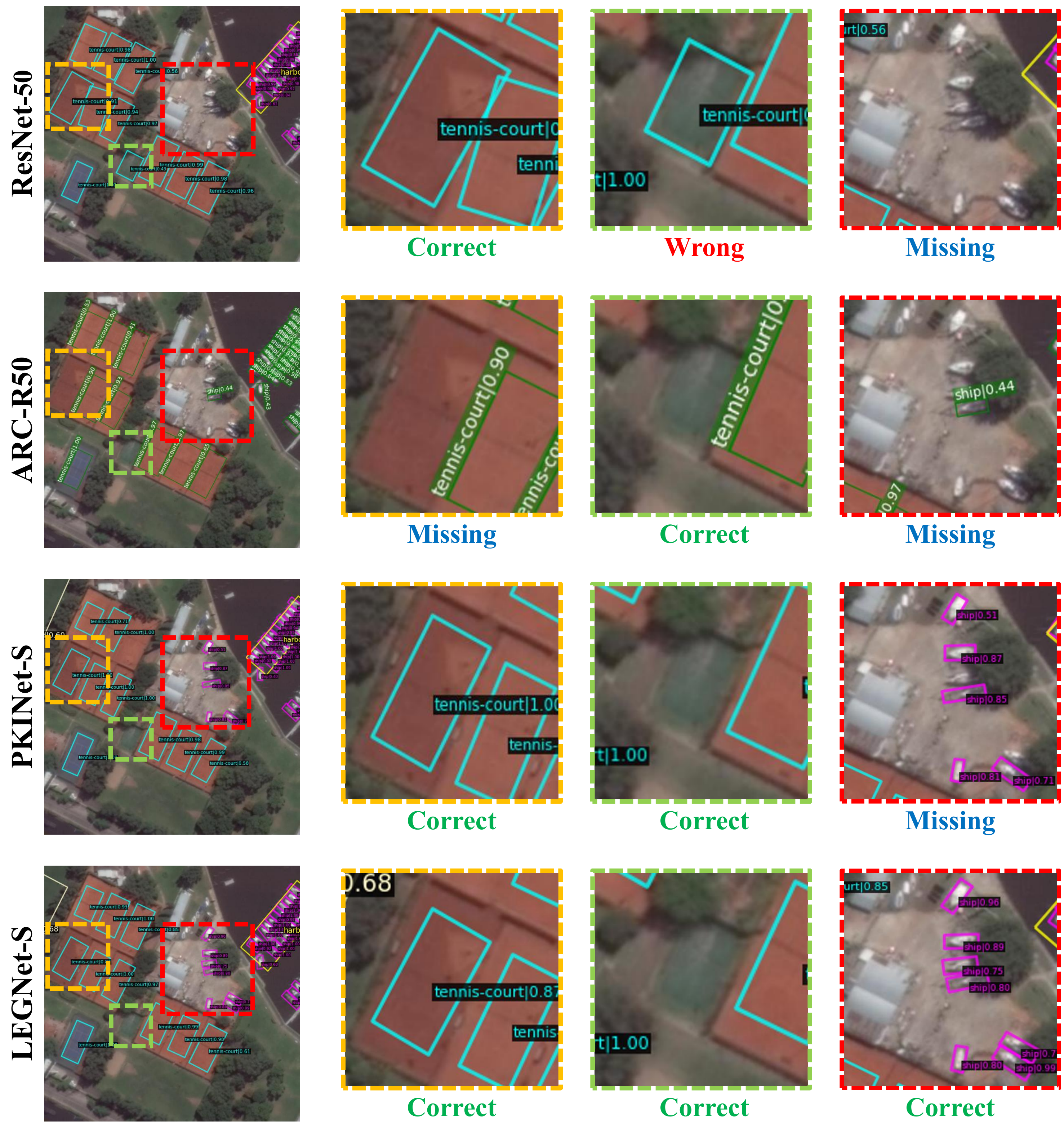}
		\end{center}
		\caption{Visualization of detection results on DOTA-v1.0 test set. Input images resolution were 1,024 $\times$ 1,024.}	\vspace{-2mm}
		\label{vis2}
	\end{figure*}

\begin{figure*}[hb]
		\begin{center}
			\includegraphics[width=1\linewidth]{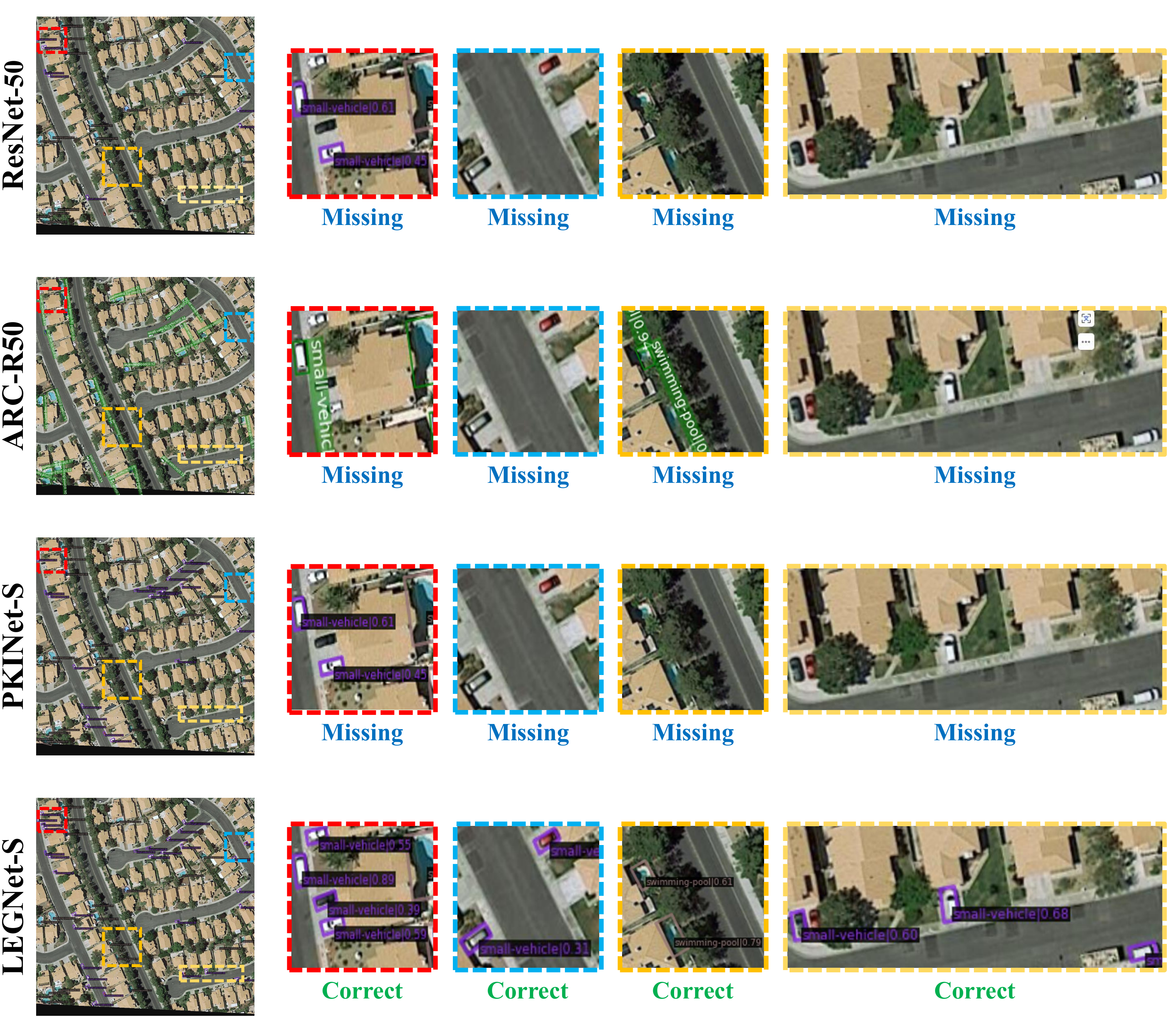}
		\end{center}
		\caption{Visualization of detection results on DOTA-v1.0 test set. Input images resolution were 1,024 $\times$ 1,024.}	\vspace{-2mm}
		\label{vis5}
	\end{figure*}

\begin{figure*}[hb]
		\begin{center}
			\includegraphics[width=1\linewidth]{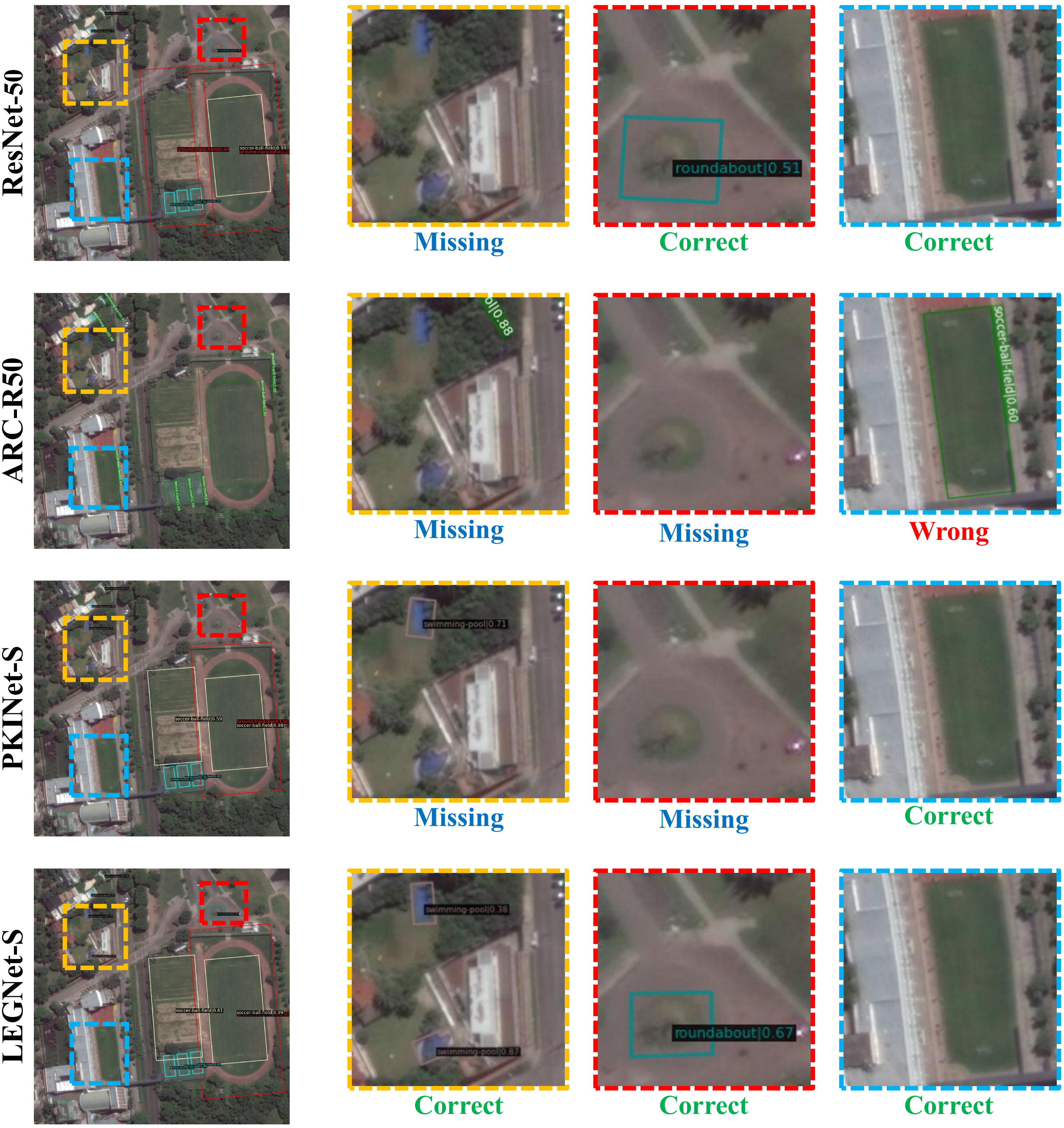}
		\end{center}
		\caption{Visualization of detection results on DOTA-v1.0 test set. Input images resolution were 1,024 $\times$ 1,024.}	\vspace{-2mm}
		\label{vis3}
	\end{figure*}

\begin{figure*}[hb]
		\begin{center}
			\includegraphics[width=1\linewidth]{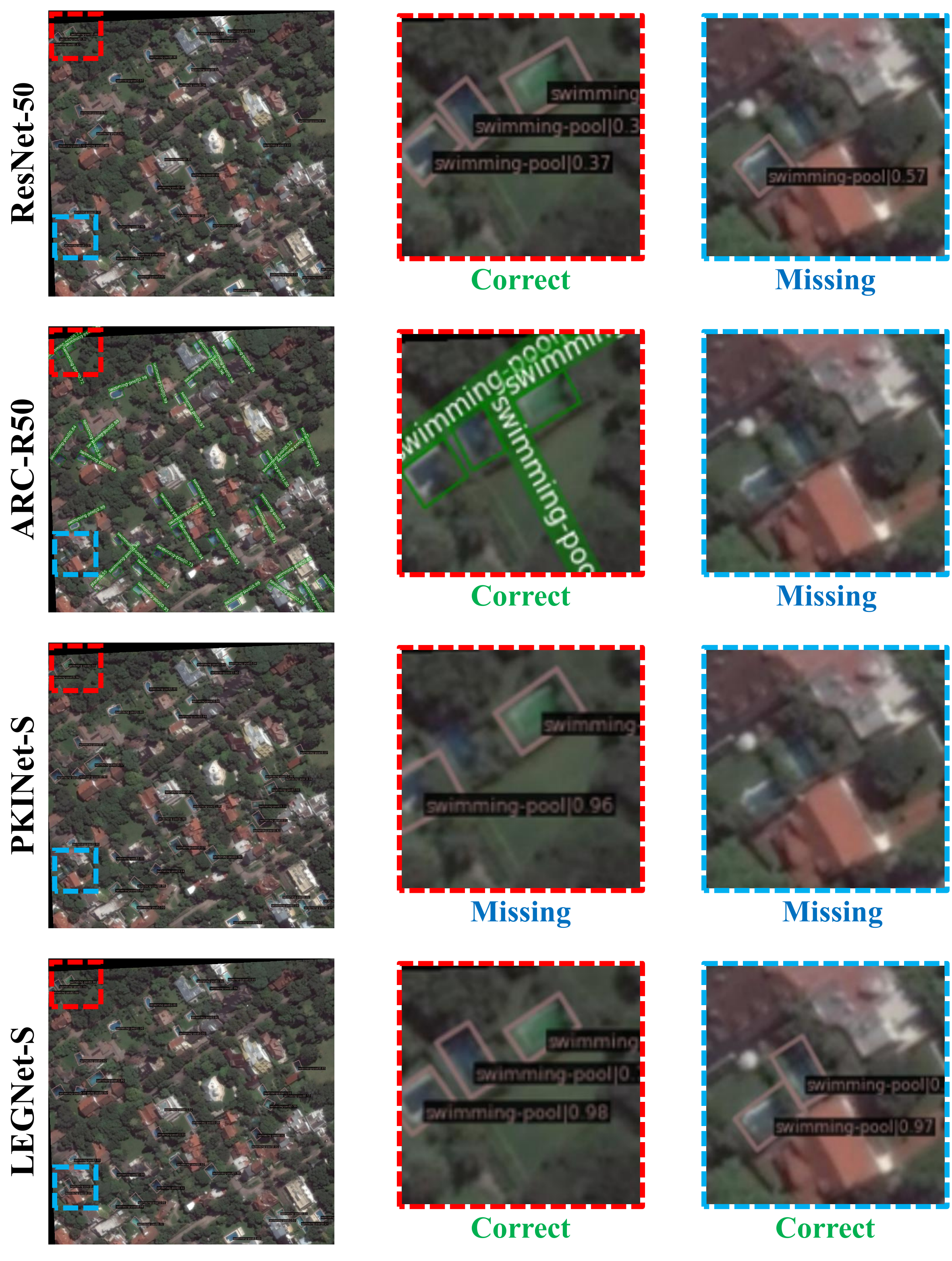}
		\end{center}
		\caption{Visualization of detection results on DOTA-v1.0 test set. Input images resolution were 1,024 $\times$ 1,024.}	\vspace{-2mm}
		\label{vis1}
	\end{figure*}

	The visual results on the DOTA-v1.0 test set are presented in~\cref{vis4,vis2,vis5,vis3,vis1}. We conducted a visual analysis using SOTA backbones, including ARC-R50~\cite{pu2023adaptive}  and PKINet~\cite{cai2024pkinet}, both of which were designed specifically for RSOD tasks. The visualization results of ARC-R50 were based on the MMDetection toolbox, while ResNet-50~\cite{he2016deep}, PKINet-S and LEGNet-S visualization results were based on the MMRotate toolbox. All models were developed based on the O-RCNN \cite{xie2021oriented} detector.
	
\end{document}

%% file: sec/0_abstract.tex
\begin{abstract}
Remote sensing object detection (RSOD) often suffers from degradations such as low spatial resolution, sensor noise, motion blur, and adverse illumination. These factors diminish feature distinctiveness, leading to ambiguous object representations and inadequate foreground-background separation. Existing RSOD methods exhibit limitations in robust detection of low-quality objects. To address these pressing challenges, we introduce LEGNet, a lightweight backbone network featuring a novel Edge-Gaussian Aggregation (EGA) module specifically engineered to enhance feature representation derived from low-quality remote sensing images. EGA module integrates: (a) orientation-aware Scharr filters to sharpen crucial edge details often lost in low-contrast or blurred objects, and (b) Gaussian-prior-based feature refinement to suppress noise and regularize ambiguous feature responses, enhancing foreground saliency under challenging conditions. EGA module alleviates prevalent problems in reduced contrast, structural discontinuities, and ambiguous feature responses prevalent in degraded images, effectively improving model robustness while maintaining computational efficiency. Comprehensive evaluations across five benchmarks (DOTA-v1.0, v1.5, DIOR-R, FAIR1M-v1.0, and VisDrone2019) demonstrate that LEGNet achieves state-of-the-art performance, particularly in detecting low-quality objects.
The code is available at \href{https://github.com/AeroVILab-AHU/LEGNet}{here}.
\end{abstract}

%% file: sec/1_intro.tex
\section{Introduction}	\label{secintro}

\begin{figure}[!t]	\centering
\includegraphics[width=0.92\linewidth]{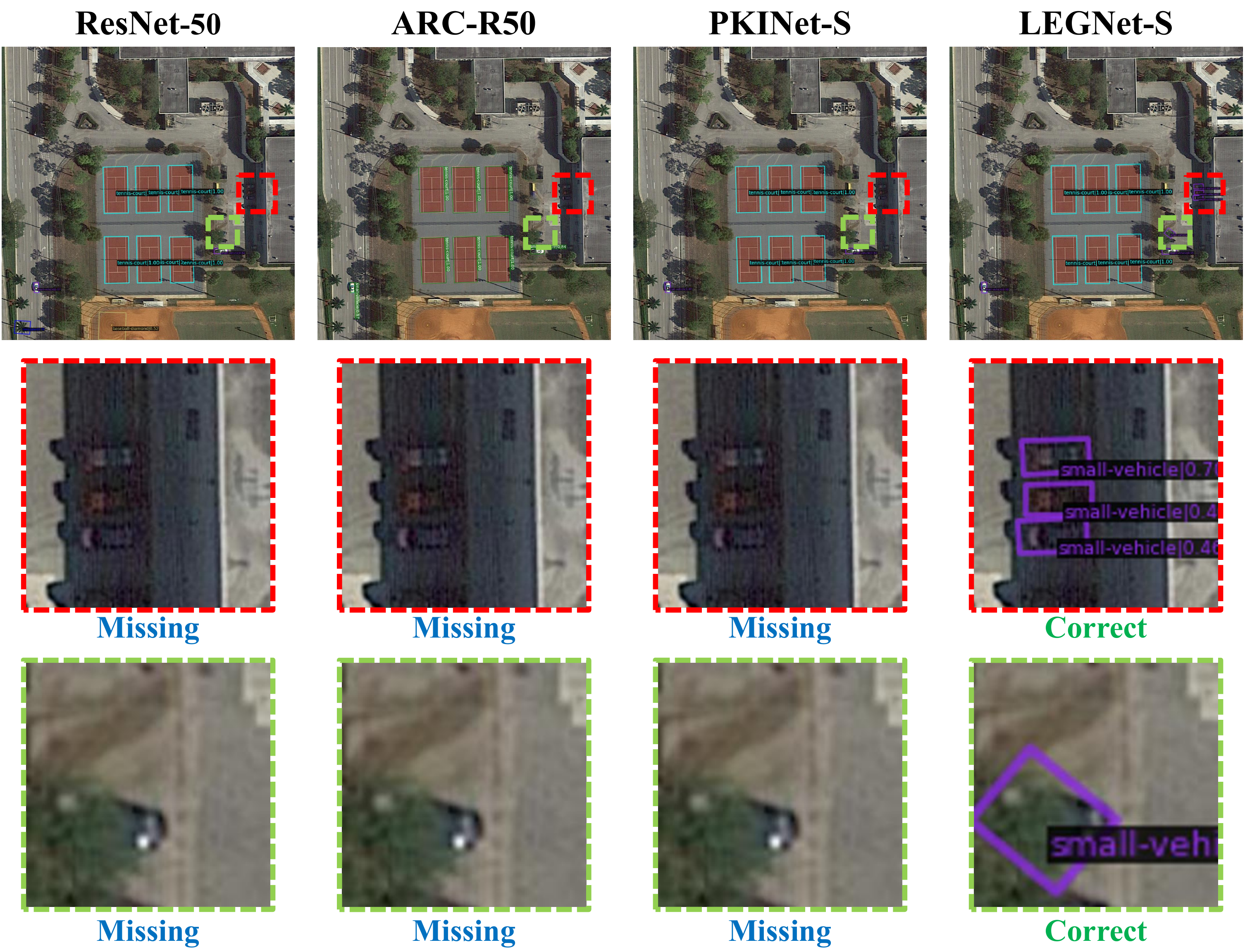}\vspace{-5pt}	
\caption{Visualization results on the DOTA-v1.0 test set~\cite{xia2018dota}. All models were built with Oriented R-CNN~\cite{xie2021oriented} detector. Our LEGNet demonstrates robust detection under challenging conditions such as occlusion (e.g., objects obscured by trees) and low-light (e.g., building shadows), surpassing previous state-of-the-art methods in both accuracy and robustness for low-quality objects.} \vspace{-4pt}	
\label{LEGNet_fig1}\vspace{-6pt}	
\end{figure} 

Remote sensing object detection (RSOD) is pivotal for diverse real-world applications, yet it contends with persistent challenges from the inherent complexities of aerial and satellite images. A critical challenge is the prevalence of low-quality objects, factors including limited sensor resolution, atmospheric interference, motion blur, variable illumination (e.g., shadows, low-light), and occlusions by natural or man-made structures. These degradation factors compromise feature discriminability, leading to: (1) diminished contrast, making foreground objects difficult to distinguish from complex backgrounds; (2) fragmented structural details and edge information, hindering precise localization; and (3) ambiguous or diminished feature responses, particularly when objects are partially obscured or under challenging lighting. Consequently, the robustness and deployability of detection models is undermined, especially for these challenging-to-discern, low-quality objects.

Numerous approaches have been developed to enhance RSOD performance, particularly for multi-scale object detection. For instance, methods employing large kernel convolutions (e.g., LSKNet~\cite{li2023large}) or multi-scale feature extraction strategies (e.g., PKINet~\cite{cai2024pkinet}) have shown promise in capturing contextual information and handling object size variations. However, despite their success in multi-scale detection, these methods often struggle to robustly represent features from low-quality or degraded objects. Consequently, the subtle yet crucial details of such low-quality objects are frequently lost, distorted, or overwhelmed by noise during feature propagation in deep networks. This problem is exacerbated in lightweight architectures, which are increasingly important for deployment on resource-constrained platforms (e.g., UAVs, satellites) but often sacrifice representational capacity for efficiency, further exhibiting difficulty in preserving subtle characteristics of degraded objects. This performance gap underscores the pressing need for lightweight networks that can specifically enhance features from low-quality RS images without incurring prohibitive computational costs.

To address this critical deficiency, we introduce LEGNet, a Lightweight Edge-Gaussian Network, designed to enhance the representation of low-quality objects in RS images while maintaining model efficiency. At the core of LEGNet is our novel Edge-Gaussian Aggregation (EGA) module. As visualized in \cref{LEGNet_fig1}, LEGNet exhibits robust performance in detecting challenging objects under conditions like occlusions and low-light environments, outperforming existing state-of-the-art (SOTA) methods, especially for objects with compromised visual quality.

The EGA module combines classical image processing techniques with modern deep learning techniques to specifically address feature degradation. It comprises two key components:
First, an orientation-aware edge enhancement mechanism leveraging Scharr operators. These operators preserve fine edge structures with rotational invariance, which is critical for identifying objects with degraded boundaries caused by resolution limits, motion, or weather effects. This explicit edge-awareness helps counteract the structural discontinuity issue.
Second, a Gaussian-prior-based feature modeling component. This part refines features, particularly those with low confidence or high ambiguity, by applying Gaussian smoothing with fixed-parameter convolutional kernels. This not only aids in attenuating noise and irrelevant background clutter but also enhances the saliency of true foreground objects, especially under varying illumination or partial occlusions, thereby addressing ambiguous feature responses and improving contrast.
The EGA module, by integrating these complementary strategies, effectively bridges traditional image processing insights with learnable deep representations. LEGNet provides a conceptually sound framework for robust feature extraction from degraded visual data, which is well-suited for RSOD tasks in challenging environments.

To validate the efficacy and efficiency of LEGNet, comprehensive experiments are conducted on four challenging public RSOD benchmarks (DOTA-v1.0~\cite{xia2018dota}, DOTA-v1.5~\cite{xia2018dota}, DIOR-R~\cite{AOPG}, FAIR1M-v1.0~\cite{sun2022fair1m}) and a UAV-view dataset (VisDrone2019~\cite{du2019visdronedet}). LEGNet consistently achieves SOTA performance across all datasets while maintaining a lightweight architecture. Specifically, it attains an mAP of 80.03\% on DOTA-v1.0, 72.89\% on DOTA-v1.5, 68.40\% on DIOR-R, 48.35\% on FAIR1M-v1.0, and 55.0\% mAP50 on VisDrone2019. These results demonstrate that LEGNet improves RSOD performance while remaining computationally efficient, making it suitable for practical, resource-constrained RS applications.

The main contributions are summarized as follows:
\begin{itemize}
	\item We introduce the novel EGA module, which combines traditional image processing operators (orientation-aware Scharr edge enhancement and Gaussian-prior-based feature modeling) with learnable deep features to specifically tackle feature degradation in low-quality RS images.
	\item We propose LEGNet, a lightweight network built upon the EGA module, designed to efficiently and effectively improve the detection of challenging objects (e.g., low-quality, blurred, or occluded) while maintaining computational efficiency suitable for edge deployment.
	\item Extensive experiments on five challenging RSOD benchmarks demonstrate that LEGNet establishes new SOTA performance, particularly excelling on low-quality objects, validating its effectiveness and practical applicability for real-world, resource-constrained RS scenarios.
\end{itemize}

%% file: sec/2_rework.tex
\section{Related Work} \label{RelatedWork}

\subsection{Oriented Object Detection Methods}
Horizontal Bounding Box (HBB) methods often struggle in RS scenarios due to their inability to represent object rotations accurately. To address this, Oriented Bounding Box (OBB) detectors have been proposed. Early methods like RRPN~\cite{yang2018automatic} introduced rotated anchors to improve object coverage. Subsequent approaches such as RoI Transformer~\cite{ding2019learning}, R$^3$Det~\cite{yang2021r3det}, and S$^2$ANet~\cite{han2021align} refined feature alignment and proposal generation. Gliding Vertex~\cite{xu2020gliding} and Oriented RepPoints~\cite{li2022oriented} further enhanced geometric flexibility. To tackle the challenges of angular periodicity and regression discontinuity, CSL~\cite{yang2020arbitrary} reformulated angle prediction as classification. Methods like SCRDet~\cite{yang2019scrdet} introduced smooth IoU-based losses. More recently, Gaussian-based metrics such as GWD~\cite{yang2021rethinking}, KLD~\cite{yang2021learning}, and KFIoU~\cite{yang2023kfiou} have provided differentiable formulations for skewed IoU landscapes, improving optimization stability. Transformer-based methods have also gained traction in RSOD. AO2-DETR~\cite{dai2023aodetr} proposed oriented proposals to enhance feature interaction, while FPNformer~\cite{tian2024fpnformer} combined FPNs with Transformer decoders to capture multi-scale and rotation-aware features.

\begin{figure*}[!t]	\centering
	\includegraphics[width=0.94\linewidth]{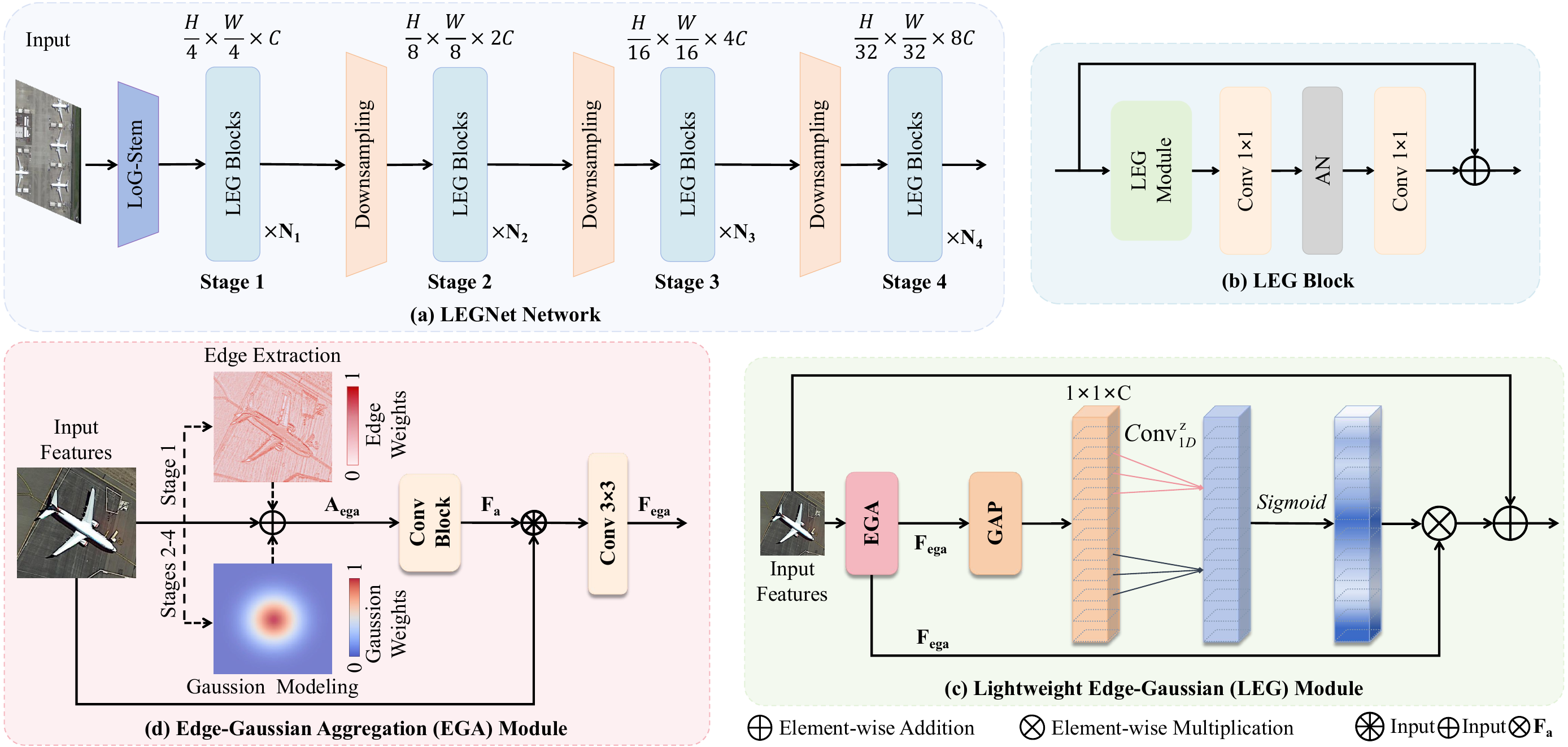}\vspace{-2pt}	
	\caption{Overview of LEGNet architecture. It consists of 4 stages with input resolutions downsampled by factors of 4, 8, 16, and 32. AN, Conv, GAP, and \( z \) denote the activation–normalization layer, convolutional layer, global average pooling, and 1D Conv size, respectively.}\label{fig_arch}
\end{figure*}

\subsection{Backbone Networks for RSOD}
Backbone design plays a critical role in RSOD performance. Recent efforts include ARC~\cite{pu2023adaptive}, which employed adaptive rotation convolutions, and LSKNet~\cite{li2023large}, which leveraged large kernels to expand receptive fields. PKINet~\cite{cai2024pkinet} integrated multi-scale kernels, while DecoupleNet~\cite{lu2024decouplenet} focused on preserving fine object details. LWGANet~\cite{lu2025lwganet} demonstrated competitive results through efficient multi-scale attention within a lightweight architecture.

Although recent methods have effectively addressed challenges related to scale, density, and orientation, RSOD continues to be hindered by degraded feature quality due to sensor limitations and adverse imaging conditions. Nevertheless, robust representation learning under such degradations remains insufficiently explored.

To address this challenge, we propose LEGNet—a lightweight edge-Gaussian network tailored for low-quality RSOD. To the best of our knowledge, LEGNet is the first RSOD backbone to integrate traditional image processing techniques (edge detection and Gaussian filtering) into a modern deep learning framework, offering a novel approach to robust feature extraction in degraded scenarios.

%% file: sec/3methods.tex
\section{Method} \label{method}
This section introduces the architecture and core components of the proposed LEGNet backbone network, which addresses RSOD challenges in low-quality images by leveraging edge cues and Gaussian modeling.

\subsection{LEGNet Architecture Overview}

As shown in \cref{fig_arch}(a), LEGNet adopts a four-stage architecture that progressively extracts multi-scale features with downsampling ratios of 1/4, 1/8, 1/16, and 1/32. It comprises three main components: a LoG-Stem layer, downsampling based on DRFD module~\cite{lu2023robust}, and lightweight edge-Gaussian (LEG) blocks.

The input image \(\mathbf{I} \in \mathbb{R}^{H \times W \times 3}\) first passes through the LoG-Stem layer, where a Laplacian of Gaussian (LoG) filter~\cite{kong2013generalized} performs initial downsampling while capturing edge features. This produces a feature map \(\mathbf{F}_{\text{stem}} \in \mathbb{R}^{\frac{H}{4} \times \frac{W}{4} \times C}\), where \(C\) set to 32 for LEGNet-Tiny and 64 for LEGNet-Small. Stage 1 processes \(\mathbf{F}_{\text{stem}}\) using \(N_1\) LEG blocks to refine 1/4-scale features. Stages 2 to 4 begin with DRFD modules, which combine convolution and max pooling to preserve small-scale features while reducing resolution. They are followed by \(N_2\), \(N_3\), and \(N_4\) LEG blocks, which further refine features at scales of 1/8, 1/16, and 1/32. LEG block enhances feature representation through either edge cues (in shallow stages) or Gaussian modeling (in deeper stages) to improve robustness against noise and shape variations. Finally, outputs from all four stages are aggregated by the RSOD decoder to generate detection results. Detailed configurations are provided in the Appendix.

\subsection{LoG-Stem Layer}
To alleviate the challenges of noisy and degraded RS images, we introduce the LoG-Stem layer as a key component of LEGNet. It enhances edge features at the initial stage by leveraging the LoG filter’s dual capability of noise suppression and edge detection—critical for RS images where the deep learning feature extractors are often inadequate.

The LoG filter combines Gaussian smoothing with a Laplacian operator, suppressing noise while highlighting regions of rapid intensity change. In our implementation, the input image \(\mathbf{I} \in \mathbb{R}^{H \times W \times 3}\) is first processed by a \(7 \times 7\) convolution to extract initial features, followed by a LoG filter with kernel size \(7 \times 7\) and \(\sigma = 1.0\), enabling edge-aware feature representation. The $2D$ LoG filter at position \(\mathbf{\textbf{x}} = (i, j)\) is defined as:
\begin{equation}
	\texttt{LoG}^{k \times k}_{\sigma}(\mathbf{\textbf{x}}) = \frac{1}{\pi \sigma^4} \left(1 - \frac{i^2 + j^2}{\sigma^2} \right) e^{-\frac{i^2 + j^2}{2\sigma^2}},
\end{equation}
with the Gaussian filter \( \texttt{G}^{k \times k}_{\sigma}(\mathbf{\textbf{x}}) \) given by:
\begin{equation}\label{gaussian_ke}
	\texttt{G}^{k \times k}_{\sigma}(\mathbf{\textbf{x}}) = \frac{1}{2\pi\sigma^2} e^{-\frac{i^2 + j^2}{2\sigma^2}},
\end{equation} 
where \(k \times k\) denotes the kernel size, and \(\sigma\) is the standard deviation, set to 1.0 in our case for optimal edge detection.

The output of the LoG filter is activated $+$ normalized (AN), and added to the input via a residual connection:
$\mathbf{F}_{LoG} = \texttt{Norm}\left(\mathbf{I} + \texttt{AN}\left(\texttt{LoG}^{7 \times 7}_{1.0}\left(\texttt{Conv}^{7 \times 7}_{2D}(\mathbf{I})\right)\right)\right).$
This residual-like design preserves image details and promotes stable gradient flow.
\(\mathbf{F}_{LoG}\) is then passed through two \(3 \times 3\) convolutional layers, with the second using stride 2 for downsampling:
$\mathbf{F}_1 = \texttt{ConvD}^{3 \times 3}_{2D}\left(\texttt{Conv}^{3 \times 3}_{2D}(\mathbf{F}_{LoG})\right),$
reducing spatial size to \(\frac{H}{2} \times \frac{W}{2}\).

To further enhance multi-scale features, \(\mathbf{F}_1\) is processed by Gaussian filters with kernel sizes \(9 \times 9\) and \(5 \times 5\), both using \(\sigma = 0.5\). The outputs are summed, normalized, and passed through the DRFD module to obtain the final 1/4-resolution feature map:
$\mathbf{F}_{LoG-Stem} = \texttt{DRFD}\left(\texttt{G}^{5\times5}_{0.5}\left(\texttt{Norm}(\texttt{G}^{9\times9}_{0.5}(\mathbf{F}_1) + \mathbf{F}_1)\right)\right).$

This process enriches edge features and integrates multi-scale context, providing a robust foundation for downstream detection. The resulting feature map \(\mathbf{F}_{LoG-Stem} \in \mathbb{R}^{\frac{H}{4} \times \frac{W}{4} \times C}\) is fed into the Stage 1 LEG blocks.

\subsection{LEG Block}    
As illustrated in \cref{fig_arch}(b), the LEG block enhances low-quality features via the EGA module. First, we apply the LEG module to the input feature $\mathbf{F}_{in}$, then a \(1 \times 1\) convolution expands channels from $C$ to $2C$, followed by an activation–normalization (AN) layer to yield
$\mathbf{F}_{mid} = \mathrm{AN}\bigl(\mathrm{Conv}^{1\times1}_{2D}(\mathrm{LEG}(\mathbf{F}_{in}))\bigr).$
Next, a second \(1 \times 1\) convolution reduces channels back to $C$, followed by normalization and dropout (rate 0.1). Its output is added to $\mathbf{F}_{in}$ to produce the final output
$\mathbf{F}_o = \mathbf{F}_{in} + \mathrm{Norm}\bigl(\mathrm{Drop}(\mathrm{Conv}^{1\times1}_{2D}(\mathbf{F}_{mid}))\bigr).$

\subsubsection{Edge Extraction}
At Stage 1 (feature size 1/4 of the input), edge and high-frequency details remain salient. We use the Scharr filter—an improved Sobel variant with better rotation invariance—to extract robust edges. Its kernels are
\[
S_x = \begin{bmatrix} -3 & 0 & 3 \\ -10 & 0 & 10 \\ -3 & 0 & 3 \end{bmatrix},\quad
S_y = \begin{bmatrix} -3 & -10 & -3 \\ 0 & 0 & 0 \\ 3 & 10 & 3 \end{bmatrix},
\]
where \( S_x \) and \( S_y \) denote the parameters of the respective convolution filters.
We fix these weights in two convolutions $\mathrm{Conv}^{S_x}_{2D}$ and $\mathrm{Conv}^{S_y}_{2D}$, then combine their outputs via
$\mathbf{A}_{edge} = \sqrt{\bigl(\mathrm{Conv}^{S_x}_{2D}(\mathbf{F}_{in})\bigr)^2 + \bigl(\mathrm{Conv}^{S_y}_{2D}(\mathbf{F}_{in})\bigr)^2}.
$

\subsubsection{Gaussian Modeling} \label{Gaussian_Modeling}
In Stages 2-4, edges are blurred with decreasing resolution, but object features  tend to exhibit a Gaussian distribution centered around a specific point. To better capture these object features, Gaussian convolution kernels based on a Gaussian spatial prior are employed, which help to smooth feature variations and suppress noise, thereby preserving critical object structures and preventing their degradation during the forward propagation of the network. As described in \cref{gaussian_ke}, the Gaussian filter assigns higher weights to pixels closer to the center of the kernel, emphasizing prominent features in the image while suppressing background noise. This is effective for stabilizing feature representations, enabling the network to focus on the most informative regions, and minimizing artifacts from high-frequency noise.
We apply a fixed \(5 \times 5\) Gaussian kernel via depthwise convolution:
$\mathbf{A}_{gauss} = G^{5\times5}_{1.0}(\mathbf{F}_{in}),$
which smooths each channel, emphasizing central pixels and suppressing noise.

\subsubsection{LEG Module}

As shown in \cref{fig_arch}(c), LEG module processes the input feature \(\mathbf{F}_{in}\) through an EGA submodule, which yield the features \( \mathbf{F}_{ega} \). To emphasize critical channels, we adopt the ECA strategy~\cite{wang2020ecanet}. The ECA is represented as:
\[
\mathbf{F}_{temp} = \text{Sigmoid}(\text{Conv}^{z}_{1D}(\text{GAP}(\mathbf{F}_{ega}))),
\]
\[
\mathbf{F}_o = \text{Norm}((\mathbf{F}_{temp} \otimes \mathbf{F}_{ega}) + \mathbf{F}_{in}),
\]
where \( \texttt{Conv}^{z}_{1D} \)  is an adaptive one-dimensional convolution with kernel size \(z\) proportional to the channels \(C\), GAP is channel-wise global average pooling, and \(\otimes\) indicates element-wise multiplication. 

\subsubsection{EGA Module}
As shown in \cref{fig_arch}(d), EGA module processes the input \(\mathbf{F}_{in}\) through a selection mechanism, which applies either edge extraction or Gaussian modeling depending on the stage:
\[
\mathbf{A}_{ega} = \begin{cases}
	\mathbf{A}_{edge}(\mathbf{F}_{in}), & \text{if Stage = 1}, \\
	\mathbf{A}_{gauss}(\mathbf{F}_{in}), & \text{otherwise}.
\end{cases}
\]
we add $\mathbf{A}_{ega}$ to $\mathbf{F}_{in}$ and pass the sum through a three-layer Conv\_Block to enhance feature representation:
\[
\mathbf{F}_{temp} = \mathrm{Conv}^{3\times3}_{2D}(\mathrm{AN}(\mathrm{Conv}^{1\times1}_{2D}(\mathbf{F}_{in} + \mathbf{A}_{ega}(\mathbf{F}_{in})))),\]
\[
\mathrm{Conv\_Block}(\mathbf{F}_{in})=\mathrm{Norm}\bigl(\mathrm{Conv}^{1\times1}_{2D}(\mathrm{AN}(\mathbf{F}_{temp}))\bigr).
\]

The Conv\_Block result \(\mathbf{F}_a\) combines \(\mathbf{F}_{in}\) using element-wise multiplication and addition operations, followed by a \(3 \times 3\) convolution:
$\mathbf{F}_{ega} = \text{Conv}^{3\times3}_{2D}((\mathbf{F}_{in} \otimes \mathbf{F}_a) + \mathbf{F}_{in})$.

\subsection{Macro Designs}
Our fine-tuning experiments demonstrated that edge information is most effective when integrated into the shallow layers of the network. This effectiveness can be attributed to the fact that shallow layers are primarily tasked with capturing low-level, high-frequency visual cues such as edges, textures, and local patterns, which are crucial for defining object boundaries—particularly in degraded images where such details are subtle. However, the integration of these low-level edge features into deeper layers has a detrimental effect on performance. This is typically attributed to the introduction of edge interferences, which causes feature redundancy and disrupts the semantic hierarchy in deeper layers, thereby leading to poor convergence and training instability. In contrast, in deeper layers, explicit edge extraction is omitted to avoid redundancy, and Gaussian modeling is applied to deep features, which are represented as Gaussian distributions—an approach that better aligns with the abstract nature of deeper representations.

To reduce noise in RS images, the LoG-Stem module is applied in shallow layers, allowing edge features to be enhanced by suppressing noise interference. 
To facilitate the effective fusion of information from both the original input and its corresponding edge maps, the LEG module is designed with extensive residual connections. These connections ensure a balanced integration of both components. Additionally, several standard convolutional layers are employed to further refine the fused features.


%% file: sec/4exp.tex
\begin{table*}[ht]
\setlength{\tabcolsep}{2pt}
\renewcommand\arraystretch{1.2}
\scriptsize      
\centering
\resizebox{\textwidth}{!}{
\hspace{-1.0em}
\begin{tabular}{r|c||c|ccccccccccccccc||c} \specialrule{0.8pt}{0pt}{0pt}
\rowcolor[rgb]{0.92,0.92,0.92} \textbf{Method}  
& \textbf{Backbone}  & \#\textbf{P} $\downarrow$  
&PL &BD &BR &GTF &SV &LV &SH &TC &BC &ST &SBF &RA &HA &SP &HC &\textbf{mAP} $\uparrow$ \\ \hline \hline
\multicolumn{19}{l}{\textbf{\textit{One-stage}}}\\	\hline
R$^3$Det~\cite{yang2021r3det} & ResNet-50~\cite{he2016deep} & 41.9M & 89.00 & 75.60 & 46.64 & 67.09 & 76.18 & 73.40 & 79.02 & 90.88 & 78.62 & 84.88 & 59.00 & 61.16 & 63.65 & 62.39 & 37.94 & 69.70 \\	\hline
SASM~\cite{hou2022shape}  & ResNet-50~\cite{he2016deep} & 36.6M & 86.42 & 78.97 & 52.47 & 69.84 & 77.30 & 75.99 & 86.72 & \textcolor{blue}{90.89} & 82.63 & 85.66 & 60.13 & 68.25 & 73.98 & 72.22 & 62.37 & 74.92 \\ \hline
O-RepPoints~\cite{li2022oriented}  & ResNet-50~\cite{he2016deep} & 36.6M & 87.02 & 83.17 & 54.13 & 71.16 & 80.18 & 78.40 & 87.28 & \textcolor{red}{90.90} & 85.97 & 86.25 & 59.90 & \textcolor{blue}{70.49} & 73.53 & 72.27 & 58.97 & 75.97 \\ \hline
R$^3$Det-GWD~\cite{yang2021rethinking} & ResNet-50~\cite{he2016deep} & 41.9M & 88.82 & 82.94 & 55.63 & 72.75 & 78.52 & 83.10 & 87.46 & 90.21 & 86.36 & 85.44 & 64.70 & 61.41 & 73.46 & 76.94 & 57.38 & 76.34 \\ \hline
R$^3$Det-KLD~\cite{yang2021learning} & ResNet-50~\cite{he2016deep} & 41.9M & 88.90 & 84.17 & 55.80 & 69.35 & 78.72 & 84.08 & 87.00 & 89.75 & 84.32 & 85.73 & 64.74 & 61.80 & 76.62 & 78.49 & \textcolor{blue}{70.89} & 77.36 \\ \hline
\multirow{4}{*}{S$^2$ANet~\cite{han2021align}} & ResNet-50~\cite{he2016deep} & 38.5M & 89.11 & 82.84 & 48.37 & 71.11 & 78.11 & 78.39 & 87.25 & 90.83 & 84.90 & 85.64 & 60.36 & 62.60 & 65.26 & 69.13 & 57.94 & 74.12  \\
& ARC-R50~\cite{pu2023adaptive} & 71.8M & 89.28 & 78.77 & 53.00 & 72.44 & 79.81 & 77.84 & 86.81 & 90.88 & 84.27 & 86.20 & 60.74 & 68.97 & 66.35 & 71.25 & 65.77 & 75.49\\
& PKINet-S~\cite{cai2024pkinet} & 24.8M & 89.67 & 84.16 & 51.94 & 71.89 & 80.81 & 83.47 & 88.29 & 90.80 & 87.01 & \textcolor{blue}{86.94} & 65.02 & 69.53 & 75.83 & 80.20 & 61.85 & 77.83 \\
\rowcolor[rgb]{0.9,0.9,0.9}&\textbf{LEGNet-S (Ours)} &23.9M
& 89.22	& 85.04	& 55.59	& 75.73	& \textcolor{red}{81.42}	& 84.56	& 88.32	& \textcolor{red}{90.90}	& \textcolor{blue}{87.90}	& \textcolor{red}{87.02}	& 65.98	& 70.13	& 77.56	& \textcolor{blue}{81.88}	& 69.26	& \textcolor{blue}{79.37} \\ \hline	

\multicolumn{19}{l}{\textbf{\textit{Two-stage}}} \\	\hline
CenterMap~\cite{long2021creating} & ResNet-50~\cite{he2016deep} & 41.1M & 89.02 & 80.56 & 49.41 & 61.98 & 77.99 & 74.19 & 83.74 & 89.44 & 78.01 & 83.52 & 47.64 & 65.93 & 63.68 & 67.07 & 61.59 & 71.59 \\ \hline
SCRDet~\cite{yang2019scrdet} & ResNet-50~\cite{he2016deep} & 41.9M & \textcolor{red}{89.98} & 80.65 & 52.09 & 68.36 & 68.36 & 60.32 & 72.41 & 90.85 & 87.94 & 86.86 & 65.02 & 66.68 & 66.25 & 68.24 & 65.21 & 72.61 \\ \hline
FR-O~\cite{ren2016faster} & ResNet-50~\cite{he2016deep} & 41.1M & 89.40 & 81.81 & 47.28 & 67.44 & 73.96 & 73.12 & 85.03 & \textcolor{red}{90.90} & 85.15 & 84.90 & 56.60 & 64.77 & 64.70 & 70.28 & 62.22 & 73.17  \\ \hline
Roi Trans.~\cite{ding2019learning} & ResNet-50~\cite{he2016deep} & 55.1M & 89.01 & 77.48 & 51.64 & 72.07 & 74.43 & 77.55 & 87.76 & 90.81 & 79.71 & 85.27 & 58.36 & 64.11 & 76.50 & 71.99 & 54.06 &  74.05 \\	\hline
G.V.~\cite{xu2020gliding}  & ResNet-50~\cite{he2016deep} & 41.1M & 89.64 & 85.00 & 52.26 & 77.34 & 73.01 & 73.14 & 86.82 & 90.74 & 79.02 & 86.81 & 59.55 & \textcolor{red}{70.91} & 72.94 & 70.86 & 57.32 & 75.02 \\ \hline
ReDet~\cite{han2021redet}  & ResNet-50~\cite{he2016deep} & 31.6M & 88.79 & 82.64 & 53.97 & 74.00 & 78.13 & 84.06 & 88.04 & \textcolor{blue}{90.89} & 87.78 & 85.75 & 61.76 & 60.39 & 75.96 & 68.07 & 63.59 & 76.25 \\ \hline
\multirow{7}{*}{O-RCNN~\cite{xie2021oriented}} & ResNet-50~\cite{he2016deep} & 41.1M & 89.46 & 82.12 & 54.78 & 70.86 & 78.93 & 83.00 & 88.20 & \textcolor{red}{90.90} & 87.50 & 84.68 & 63.97 & 67.69 & 74.94 & 68.84 & 52.28 & 75.87 \\
& ARC-R50~\cite{pu2023adaptive} & 74.4M & 89.40 & 82.48 & 55.33 & 73.88 & 79.37 & 84.05 & 88.06 & \textcolor{red}{90.90} & 86.44 & 84.83 & 63.63 & 70.32 & 74.29 & 71.91 & 65.43 &  77.35 \\
& LSKNet-S~\cite{li2023large}  & 31.0M  & 89.66 & 85.52 & \textcolor{red}{57.72} & 75.70 & 74.95 & 78.69 & 88.24 & 90.88 & 86.79 & 86.38 & \textcolor{blue}{66.92} & 63.77 & 77.77 & 74.47 & 64.82 & 77.49    \\
& DecoupleNet-D2~\cite{lu2024decouplenet}  & \textcolor{blue}{23.3M}  &89.37	&83.25	&54.29	&75.51	&79.83	&84.82	&88.49	&\textcolor{blue}{90.89}	&87.19	&86.23	&66.07	&65.53	&77.23	&72.34	&69.62 & 78.04    \\
& PKINet-S~\cite{cai2024pkinet} & 30.8M & 89.72 & 84.20 & 55.81 & \textcolor{blue}{77.63} & 80.25 & 84.45 & 88.12 & 90.88 & 87.57 & 86.07 & 66.86 & 70.23 & 77.47 & 73.62 & 62.94 & 78.39 \\
\rowcolor[rgb]{0.9,0.9,0.9}&\textbf{LEGNet-T (Ours)} & \textcolor{red}{20.6M}
& 89.45	& \textcolor{red}{86.49}	& 55.76	& 76.38	& 80.59	& \textcolor{red}{85.40}	& 88.42	& \textcolor{red}{90.90}	& \textcolor{red}{88.72}	& 86.42	& 65.24	& 67.81	& \textcolor{blue}{77.93}	& 73.49	& \textcolor{red}{71.39}	& 78.96 \\
\rowcolor[rgb]{0.9,0.9,0.9}&\textbf{LEGNet-S (Ours)} & 29.8M
& \textcolor{blue}{89.87}	& \textcolor{blue}{86.32}	& \textcolor{blue}{56.36}	& \textcolor{red}{78.67}	& \textcolor{blue}{81.34}	& \textcolor{blue}{85.29}	& \textcolor{blue}{88.53}	& \textcolor{blue}{90.89}	& 87.31 & 86.46	& \textcolor{red}{70.38}	& 68.48	& \textcolor{red}{78.34}	& \textcolor{red}{82.00}	& 70.25	& \textcolor{red}{80.03} \\ 	\hline
\end{tabular}} \vspace{-8pt}
\caption{Performance comparison on DOTA-v1.0 test set~\cite{xia2018dota} under single-scale training and testing. Results for other methods are sourced from~\cite{cai2024pkinet,lu2024decouplenet}. ``\#\textbf{P}'' is calculated for backbone + detector. \textcolor{red}{Red} and \textcolor{blue}{blue} indicate the best and second-best results per column, respectively.} \label{tab_det_dota10}\vspace{-4pt}
\end{table*}

\section{Experiments}
This section presents a comprehensive evaluation of the proposed LEGNet on the RSOD task. We conducted extensive experiments on four RSOD benchmark datasets: DOTA-v1.0 and v1.5~\cite{xia2018dota}, DIOR-R~\cite{AOPG}, and FAIR1M-v1.0~\cite{sun2022fair1m}, as well as a UAV-based dataset: VisDrone2019~\cite{du2019visdronedet}. Results on DOTA-v1.0, DOTA-v1.5, and FAIR1M-v1.0 were obtained through official online testing to ensure fair and reproducible evaluation. These datasets present diverse challenges, including scale variation, poor-quality features, fine-grained categories, and small object detection, offering a thorough test of LEGNet’s robustness and effectiveness.

\subsection{Implementation Details}
For a fair comparison with existing SOTA methods~\cite{pu2023adaptive, li2023large, cai2024pkinet}, we pre-trained our backbone on ImageNet-1K~\cite{deng2009imagenet} for 300 epochs to enhance model robustness on RSOD datasets. 

For training and evaluation, we followed the configurations from previous works~\cite{pu2023adaptive, li2023large, cai2024pkinet, lu2024decouplenet}. Specifically, for DOTA-v1.0, DOTA-v1.5, and DIOR-R, we applied single-scale training/testing with a scale factor of 3$\times$. DOTA images were cropped into 1,024 $\times$ 1,024 patches with 200-pixel overlap, while DIOR-R used 800 $\times$ 800 inputs. For FAIR1M-v1.0, we followed~\cite{li2023large} and adopted multi-scale training/testing (scales: 0.5, 1.0, 1.5) with 1,024 $\times$ 1,024 crops and 500-pixel overlap. For VisDrone2019, we used single-scale training/testing with a 1$\times$ scale and 1,333 $\times$ 800 input size. The AdamW optimizer~\cite{loshchilov2017decoupled} was used with a momentum of 0.9 and weight decay of 0.05, an initial learning rate of 0.0002. A cosine learning rate schedule~\cite{loshchilov2016sgdr} with warm-up was adopted. Random resizing and flipping were applied during training. The test stage used the same resolution as training for consistency. 

All experiments were conducted using MMRotate~\cite{zhou2022mmrotate} and PyTorch~\cite{paszke2019pytorch} on Ubuntu 20.04. We used 4 NVIDIA RTX 3090 GPUs for both pre-training and RSOD experiments. Unless otherwise specified, LEGNet was integrated into the O-RCNN~\cite{xie2021oriented} detector.

\subsection{Quantitative Results} 
\subsubsection{\textbf{Performance on DOTA-v1.0}}
LEGNet achieves SOTA performance on the DOTA-v1.0 benchmark under single-scale training and testing protocols. As shown in~\cref{tab_det_dota10}, LEGNet-S with the O-RCNN~\cite{xie2021oriented} two-stage detector achieves 80.03\% mAP (29.8M parameters), surpassing existing methods while maintaining competitive model complexity. With the S$^2$ANet~\cite{han2021align} single-stage detector, LEGNet-S (23.9M parameters) attains 79.37\% mAP. Notably, LEGNet variants secure top-two results in 13 of the 15 categories.

Compared to other lightweight models, LEGNet-T sets a new SOTA accuracy-efficiency trade-off, reaching 78.96\% mAP with only 20.6M parameters—the most lightweight configuration listed. This is a 0.92\% mAP improvement over the previous leading efficient model, DecoupleNet-D2~\cite{lu2024decouplenet}, using 11.6\% fewer parameters.

LEGNet demonstrates superior performance on challenging categories, often characterized by low-quality instances: (1) For small vehicles (SV), LEGNet-S (with S$^2$ANet) achieves 81.42\% AP, outperforming the previous best (PKINet-S~\cite{cai2024pkinet} with S$^2$ANet) by 0.61\%. (2) For swimming pool (SP) detection, LEGNet-S (with O-RCNN) obtains 82.00\% AP, surpassing competitors by at least 1.8\%. (3) For helicopter (HC) detection, LEGNet-T achieves 71.39\% AP, outperforming PKINet-S (with O-RCNN) by 8.45\% with 33.1\% fewer parameters. These results demonstrate LEGNet's strong balance of parameter efficiency and detection accuracy for oriented objects on this benchmark. To our knowledge, this is the first instance of achieving over 80\% mAP on DOTA-v1.0 (single-scale protocols), a significant advancement in RSOD. This underscores the importance of effective feature enhancement and extraction for low-quality objects.

\begin{table}[t]
\centering	\scriptsize	
\renewcommand\arraystretch{1.2}
\begin{tabular}{cc|ccc}	\specialrule{0.8pt}{0pt}{0pt}
\multicolumn{2}{c}{\multirow{2}{*}{\textbf{Framework}}}\vline 
&\multicolumn{3}{c}{\textbf{mAP (\%) $\uparrow$}}\\ \cline{3-5}
&& ResNet-50~\cite{he2016deep} & PKINet-S~\cite{cai2024pkinet}		
&\cellcolor[rgb]{0.9,0.9,0.9} \textbf{LEGNet-S} \\	\hline 
\multirow{3}{*}{\makecell[c]{\textit{One} \\ \textit{Stage}}}	
&R-FCOS~\cite{tian2020fcos}		& 72.45 & 74.86 & \cellcolor[rgb]{0.9,0.9,0.9}76.89 \\ 
&R$^3$Det~\cite{yang2021r3det} 	& 69.70	& 75.89	& \cellcolor[rgb]{0.9,0.9,0.9}77.62 \\ 
&S$^2$ANet~\cite{han2021align} 	& 74.12	& 77.83 & \cellcolor[rgb]{0.9,0.9,0.9}79.37 \\ \hline
\multirow{3}{*}{\makecell[c]{Two \\ Stage}}	
&FR-O~\cite{ren2016faster}		& 73.17	& 76.45	& \cellcolor[rgb]{0.9,0.9,0.9}77.19	\\ 
&Roi Trans.~\cite{ding2019learning}& 74.05& 77.17	& \cellcolor[rgb]{0.9,0.9,0.9}79.62 \\ 
&O-RCNN~\cite{xie2021oriented}	& 75.87	& 78.39 & \cellcolor[rgb]{0.9,0.9,0.9}80.03 \\ \hline 
\end{tabular}	\vspace{-8pt}	 	\caption{Comparison of mAP across different detectors using various backbones on the DOTA-v1.0 test set .}\vspace{-4pt} 	\label{tab_differ_frameworks}
\end{table}

As shown in \cref{tab_differ_frameworks}, LEGNet's effectiveness as a general-purpose RSOD backbone is validated through framework compatibility experiments. Integrated with several mainstream detection frameworks, LEGNet-S consistently outperforms both ResNet-50~\cite{he2016deep} and the recent PKINet-S~\cite{cai2024pkinet} across these one-stage and two-stage paradigms.

For single-stage detectors, LEGNet-S yields mAP improvements of +7.92\% (R$^3$Det~\cite{yang2021r3det}), +4.44\% (R-FCOS~\cite{tian2020fcos}), and +5.25\% (S$^2$ANet~\cite{han2021align}) over their ResNet-50. LEGNet-S, with S$^2$ANet, achieves 79.37\% mAP.

In two-stage detectors, LEGNet-S similarly demonstrates strong performance, with mAP gains of +5.57\% for RoI Transformer~\cite{ding2019learning} and +4.16\% for O-RCNN~\cite{xie2021oriented} compared to their ResNet-50 versions. Particularly, O-RCNN equipped with LEGNet-S achieves 80.03\% mAP, surpassing its PKINet-S counterpart by 1.64\% while maintaining comparable parameter efficiency.

These comparisons demonstrate that LEGNet enhances feature representation for various detection architectures, highlighting its strong compatibility and learning capacity.

\begin{table}[t]	\centering \scriptsize	
\renewcommand{\arraystretch}{1.2}
\setlength{\tabcolsep}{2.8mm}{
\begin{tabular}{rcc|cc}	\specialrule{0.8pt}{0pt}{0pt}
\multirow{2}{*}{\textbf{Backbone}}		&\multicolumn{2}{c}{\textbf{Backbone only}}&\multirow{2}{*}{\textbf{\makecell[c]{\textbf{Speed} \\ (FPS)} $\uparrow$}}
& \multirow{2}{*}{\textbf{\makecell[c]{mAP \\ (\%)}} \textbf{$\uparrow$}}\\ 
&\#\textbf{P} $\downarrow$	&\textbf{FLOPs} $\downarrow$&&	\\ \hline 
ResNet-50~\cite{he2016deep} 	& 23.3M	& 86.1G	&23.4& 75.87  \\
ARC-R50~\cite{pu2023adaptive} 	& N/A 	& N/A		&11.8& 77.35  \\
LSKNet-S~\cite{li2023large} 	& 14.4M	& 54.4G	&22.5& 77.49  \\
DecoupleNet-D2~\cite{lu2024decouplenet}	& \textcolor{blue}{6.2M}	& \textcolor{red}{23.1G}	&\textcolor{blue}{26.9}& 78.04  \\
PKINet-S~\cite{cai2024pkinet} 	& 13.7M	& 70.2G	&5.4& 78.39\\
\rowcolor[rgb]{0.9,0.9,0.9} \textbf{LEGNet-T}& \textcolor{red}{3.6M} & \textcolor{blue}{30.2G} &\textcolor{red}{28.3}& \textcolor{blue}{78.96} \\
\rowcolor[rgb]{0.9,0.9,0.9} \textbf{LEGNet-S}&12.7M & 65.4G &20.9& \textcolor{red}{80.03} \\ \hline 
\end{tabular}}	\vspace{-8pt}	
\caption{Comparison of mAP and FPS with various backbone networks using the O-RCNN~\cite{xie2021oriented} detector on the DOTA-v1.0 test set. Inference speed, with detector, is measured on a single RTX 3090 GPU. ``\#\textbf{P}'' and FLOPs are calculated for backbones only. }
\label{tab_diff_backbone_param_flops}\vspace{-4pt} 
\end{table}

\begin{table*}[t]
\centering	\scriptsize	
\setlength\tabcolsep{5pt}
\renewcommand\arraystretch{1.2}
\begin{tabular}{r||cccccccccccccccc||c} \specialrule{0.8pt}{0pt}{0pt}
\rowcolor[rgb]{0.92,0.92,0.92} \textbf{Method} & PL    & BD    & BR    & GTF   & SV    & LV    & SH    & TC    
& BC    & ST    & SBF   & RA    & HA    & SP    & HC  &CC & \textbf{mAP $\uparrow$} \\ \hline	\hline
RetinaNet-O~\cite{lin2017focal} & 71.43 & 77.64 & 42.12 & 64.65 & 44.53 & 56.79 & 73.31 & 90.84 & 76.02 & 59.96 & 46.95 & 69.24 & 59.65 & 64.52 & 48.06 & 0.83 & 59.16 \\
FR-O~\cite{ren2016faster}  & 71.89 & 74.47 & 44.45 & 59.87 & 51.28 & 68.98 & 79.37 & 90.78 & 77.38 & 67.50 & 47.75 & 69.72 & 61.22 & 65.28 & 60.47 & 1.54 & 62.00 \\
Mask R-CNN~\cite{he2017mask}  & 76.84 & 73.51 & 49.90 & 57.80 & 51.31 & 71.34 & 79.75 & 90.46 & 74.21 & 66.07 & 46.21 & 70.61 & 63.07 & 64.46 & 57.81 & 9.42 & 62.67 \\
HTC~\cite{chen2019hybrid} & 77.80 & 73.67 & 51.40 & 63.99 & 51.54 & 73.31 & 80.31 & 90.48 & 75.12 & 67.34 & 48.51 & 70.63 & 64.84 & 64.48 & 55.87 & 5.15 & 63.40 \\
ReDet~\cite{han2021redet}  & 79.20 & 82.81 & 51.92 & 71.41 & 52.38 & 75.73 & 80.92 & 90.83 & 75.81 & 68.64 & 49.29 & 72.03 & 73.36 & 70.55 & 63.33 & 11.53 & 66.86 \\
LSKNet-S~\cite{li2023large}  
& 72.05 & 84.94 & 55.41 & \textcolor{red}{74.93} & 52.42 & \textcolor{blue}{77.45} & 81.17 & 90.85 & \textcolor{blue}{79.44} & \textcolor{blue}{69.00} & 62.10 & \textcolor{blue}{73.72} & \textcolor{red}{77.49} & \textcolor{blue}{75.29} & 55.81 & \textcolor{blue}{42.19} & 70.26 \\
SOOD~\cite{xi2024structure}  
& \textcolor{blue}{80.32} & 84.41 & 52.59 & 74.77 & \textcolor{red}{58.48} & 76.90 & 86.97 & \textcolor{blue}{90.87} & 78.62 & \textcolor{red}{76.56} & 62.93 & 71.16 & 74.64 & \textcolor{red}{76.04} & 55.97 & 25.09 & 70.39 \\
PKINet-S~\cite{cai2024pkinet} & 80.31 & \textcolor{blue}{85.00} & \textcolor{blue}{55.61} & 74.38 & 52.41 & 76.85 & \textcolor{blue}{88.38} & \textcolor{blue}{90.87} & 79.04 & 68.78 & \textcolor{red}{67.47} & 72.45 & 76.24 & 74.53 & \textcolor{blue}{64.07} & 37.13 & \textcolor{blue}{71.47} \\ \hline \hline
\rowcolor[rgb]{0.9,0.9,0.9}\textbf{LEGNet-S} (\textbf{ours}) 
& \textcolor{red}{80.48}	& \textcolor{red}{85.04}	& \textcolor{red}{55.64}	& \textcolor{blue}{74.86}	& \textcolor{blue}{52.64}	& \textcolor{red}{82.16}	& \textcolor{red}{89.11}	& \textcolor{red}{90.88}	& \textcolor{red}{84.55}	& 68.91	& \textcolor{blue}{66.21}	& \textcolor{red}{74.16}	& \textcolor{blue}{77.44}	& 74.68	& \textcolor{red}{65.76}	& \textcolor{red}{43.75}	& \textcolor{red}{72.89}\\ \hline
\end{tabular} \vspace{-8pt}
\caption{Comparison with SOTA methods on the DOTA-v1.5 test set~\cite{xia2018dota} using single-scale training and testing. LEGNet-S is developed based on the O-RCNN \cite{xie2021oriented} detector. Results for other methods are sourced from PKINet~\cite{cai2024pkinet} and SOOD~\cite{xi2024structure}.}\label{Det_DOTA15} \vspace{6pt}

\setlength\tabcolsep{10pt}
\centering \scriptsize 
\renewcommand\arraystretch{1.2}
\begin{tabular}{c|c|c|c|c|c|c|c|c} 	 \specialrule{0.8pt}{0pt}{0pt}
\textbf{Method}  & G. V.~\cite{xu2020gliding} & RetinaNet~\cite{lin2017focal} & C-RCNN~\cite{cai2018cascade} & FR-O~\cite{ren2016faster}  & RoI Trans.~\cite{ding2019learning}  & O-RCNN~\cite{xie2021oriented} & LSKNet-S~\cite{li2023large} &\cellcolor[rgb]{0.9,0.9,0.9} \textbf{LEGNet-S}\\  \hline
\textbf{mAP(\%)} $\uparrow$    & 29.92    & 30.67    & 31.18  & 32.12  & 35.29  & 45.60    & \textcolor{blue}{47.87}   & \cellcolor[rgb]{0.9,0.9,0.9} \textcolor{red}{48.35}\\ \hline
\end{tabular} \vspace{-8pt}\caption{Comparison with SOTA methods on FAIR1M-v1.0 test set~\cite{sun2022fair1m}. Results for other methods are sourced from LSKNet~\cite{li2023large}.}
\label{tab_fair} \vspace{-4pt}
\end{table*}

\cref{tab_diff_backbone_param_flops} presents a comparative evaluation of LEGNet integrated with O-RCNN~\cite{xie2021oriented}. LEGNet-S achieves state-of-the-art performance with 80.03\% mAP, outperforming all baselines while maintaining moderate complexity (12.7M params, 65.4G FLOPs) and a competitive speed (20.9 FPS), which is 1.77$\times$ faster than ARC-R50~\cite{pu2023adaptive}.

The lightweight variant LEGNet-T offers the best accuracy-efficiency trade-off, reaching 78.96\% mAP with only 3.6M parameters and 30.2G FLOPs—41.9\% fewer parameters and 0.92\% higher mAP than DecoupleNet-D2~\cite{lu2024decouplenet}. It also outperforms PKINet-S by 0.57\% mAP with 73.7\% fewer parameters and 42.7\% less computation.

Overall, LEGNet provides the largest performance gain among all backbones. It reduces parameters by 84.5\% compared to ResNet-50 while retaining high detection accuracy, setting a new efficiency benchmark. These results highlight LEGNet’s strong balance between representation and efficiency, making it well-suited for real-time RSOD applications on resource-constrained devices.


\subsubsection{\textbf{Performance on DOTA-v1.5}}
As shown in \cref{Det_DOTA15}, LEGNet-S demonstrates superior performance on the DOTA-v1.5 test set under single-scale training and testing. It achieves a top mAP of 72.89\%, surpassing PKINet-S (71.47\% mAP) and SOOD~\cite{xi2024structure} (70.39\% mAP) by 1.42\% and 2.50\% mAP, respectively. Our method secures top-two AP scores in 14 of 16 categories, including key ones like large vehicle (LV: 82.16\%), ship (SH: 89.11\%), roundabout (RA: 74.16\%), and container crane (CC: 43.75\%). LEGNet-S shows 0.01–5.11\% AP improvements in different categories, indicating robust generalization across diverse objects.

\subsubsection{\textbf{Performance on FAIR1M-v1.0}}
On the FAIR1M-v1.0 benchmark (\cref{tab_fair}), LEGNet-S achieves a new SOTA mAP of 48.35\%. This result is 0.48\% mAP higher than LSKNet-S (47.87\% mAP) and shows a 2.75\% absolute mAP gain over the O-RCNN baseline (45.60\% mAP). LEGNet-S consistently outperforms various architectures, with margins of 18.43\% mAP over Gliding Vertex~\cite{xu2020gliding} (29.92\% mAP) and 13.06\% mAP over RoI Transformer~\cite{ding2019learning} (35.29\% mAP). 

\subsubsection{\textbf{Performance on DIOR-R}}
Experimental results on the DIOR-R test set demonstrate the superior efficiency and accuracy of our LEGNet-S. As summarized in \cref{Det_DIOR_R}, LEGNet-S achieves SOTA detection performance (68.40\% mAP) with the lowest model complexity among the compared methods, utilizing only 29.9M parameters and 118.1G FLOPs. Compared to the previous leading method, PKINet-S (67.03\% mAP), our LEGNet-S achieves a 1.37\% mAP improvement while reducing the parameters by 0.9M. LEGNet-S achieves these results with computational costs equivalent to PKINet-S (118.1G FLOPs) and surpasses other efficiency-focused methods, such as Oriented Rep~\cite{li2022oriented} (66.71\% mAP at 118.8G FLOPs) and LSKNet-S (65.90\% mAP at 173.6G FLOPs).

\begin{table}[t]
\centering	\scriptsize	
\renewcommand{\arraystretch}{1.2}
\setlength\tabcolsep{8pt}
\begin{tabular}{c|ccc}	\specialrule{0.8pt}{0pt}{0pt}
\textbf{Method}
& \#\textbf{P} (M) $\downarrow$
& \textbf{FLOPs} (G) $\downarrow$ 
&\textbf{mAP} \textbf{(\%)} $\uparrow$  \\ \hline 	
FR-O~\cite{ren2016faster}				& 41.1	& 134.4	& 59.54	\\
G.V.~\cite{xu2020gliding}				& 41.1	& 134.4	& 60.06	\\		
RoI Trans.~\cite{ding2019learning}		& 55.1	& 148.4	& 63.87	\\
LSKNet-S~\cite{li2023large}				& 31.0	& 173.6	& 65.90	\\ 
Oriented Rep~\cite{li2022oriented}		& 36.6	& \textcolor{blue}{118.8}	& 66.71	\\
DCFL~\cite{DCFL}						& 36.1	& -		& 66.80	\\
PKINet-S~\cite{cai2024pkinet}			& \textcolor{blue}{30.8}	& \textcolor{red}{118.1}	& \textcolor{blue}{67.03}	\\
\rowcolor[rgb]{0.9,0.9,0.9} \textbf{LEGNet-S}		& \textcolor{red}{29.9}	& \textcolor{red}{118.1}	&\textcolor{red}{68.40} \\ \hline
\end{tabular}	\vspace{-8pt}	
\caption{Experimental results on the DIOR-R test set~\cite{AOPG}. \#\textbf{P} and FLOPs were tested by 800 $\times$ 800 with backbone + detector.}\vspace{-2pt} \label{Det_DIOR_R}
\end{table}

\begin{table}[t]
\centering \scriptsize	
\renewcommand{\arraystretch}{1.2}
\setlength{\tabcolsep}{2.5mm}{
\begin{tabular}{c|cccccc}	 \specialrule{0.8pt}{0pt}{0pt}
Method	&AP	& AP$_{50}$	& AP$_{75}$	& AP$_{s}$	& AP$_{m}$	& AP$_{l}$\\ \hline
GFL V1~\cite{li2020generalized}		& 23.4	& 38.2	& 24.2	& 14.3	& 34.7	& 40.1		\\ 	
CMDNet~\cite{duan2021coarsegrained}	& 29.2	& 49.5	& 29.8	& 16.1	& 35.9	& 36.5		\\ 	
QueryDet~\cite{yang2022querydet} 	& 28.3	& 48.1	& 28.7	& -	& -	& -	\\ 	
DINO~\cite{zhang2023dino}			& 26.8	& 44.2	& 28.9	& 17.5	& 37.3	& 41.3		\\ 	
RT-DETR~\cite{zhao2024detrs}		& 15.9	& 36.5	& 10.7	& 13.8	& 28.4	& 23.1		\\ 	
BAFNet~\cite{song2025boundaryaware}	& \textcolor{blue}{30.8}	& \textcolor{blue}{52.6}	& \textcolor{blue}{30.9}	& \textcolor{blue}{22.4}	& \textcolor{blue}{41.0}	& \textcolor{blue}{43.2}		\\ \hline
\rowcolor[rgb]{0.9,0.9,0.9} \textbf{LEGNet-S}& \textcolor{red}{32.2}& \textcolor{red}{55.0}& \textcolor{red}{32.8}& \textcolor{red}{24.5}& \textcolor{red}{42.5}& \textcolor{red}{45.0}			\\ \hline
\end{tabular}}\vspace{-8pt}
\caption{Results of LEGNet compared with SOTA methods on VisDrone2019~\cite{du2019visdronedet}. LEGNet-S is built within BAFNet~\cite{song2025boundaryaware} detector. Other experimental results were referenced from BAFNet.} \label{VisDrone}	\vspace{-3pt}
\end{table}	%

\subsubsection{\textbf{Performance on VisDrone2019}}
Results on the VisDrone2019 benchmark (\cref{VisDrone}) further confirm the effectiveness of our LEGNet-S. When integrated with the BAFNet~\cite{song2025boundaryaware} detector, LEGNet-S achieves new SOTA performance across all primary metrics, reaching 32.2\% AP—a 1.4\% absolute improvement over BAFNet (30.8\% AP). Our method excels in small object detection (AP$_s$), achieving 24.5\%, which is 2.1\% higher than BAFNet, and surpasses GFL V1 (14.3\%) and DINO (17.5\%) by 10.2\% and 7.0\%, respectively. With 55.0\% AP$_{50}$ and 32.8\% AP$_{75}$, LEGNet-S outperforms BAFNet by 2.4\% and 1.9\%, respectively, indicating superior localization accuracy at multiple IoU thresholds. LEGNet-S maintains advantages across all object scales, particularly for small objects, and also improves large object detection (45.0\% AP$_{l}$ vs. BAFNet's 43.2\%).

These findings validate LEGNet-S's enhanced feature representation capabilities, especially when combined with boundary-aware decoders. This demonstrates its effectiveness for drone-captured scenarios characterized by dense small objects and complex backgrounds.

\subsection{Qualitative Results} 
\begin{figure}[t]
	\begin{center}
		\includegraphics[width=1\linewidth]{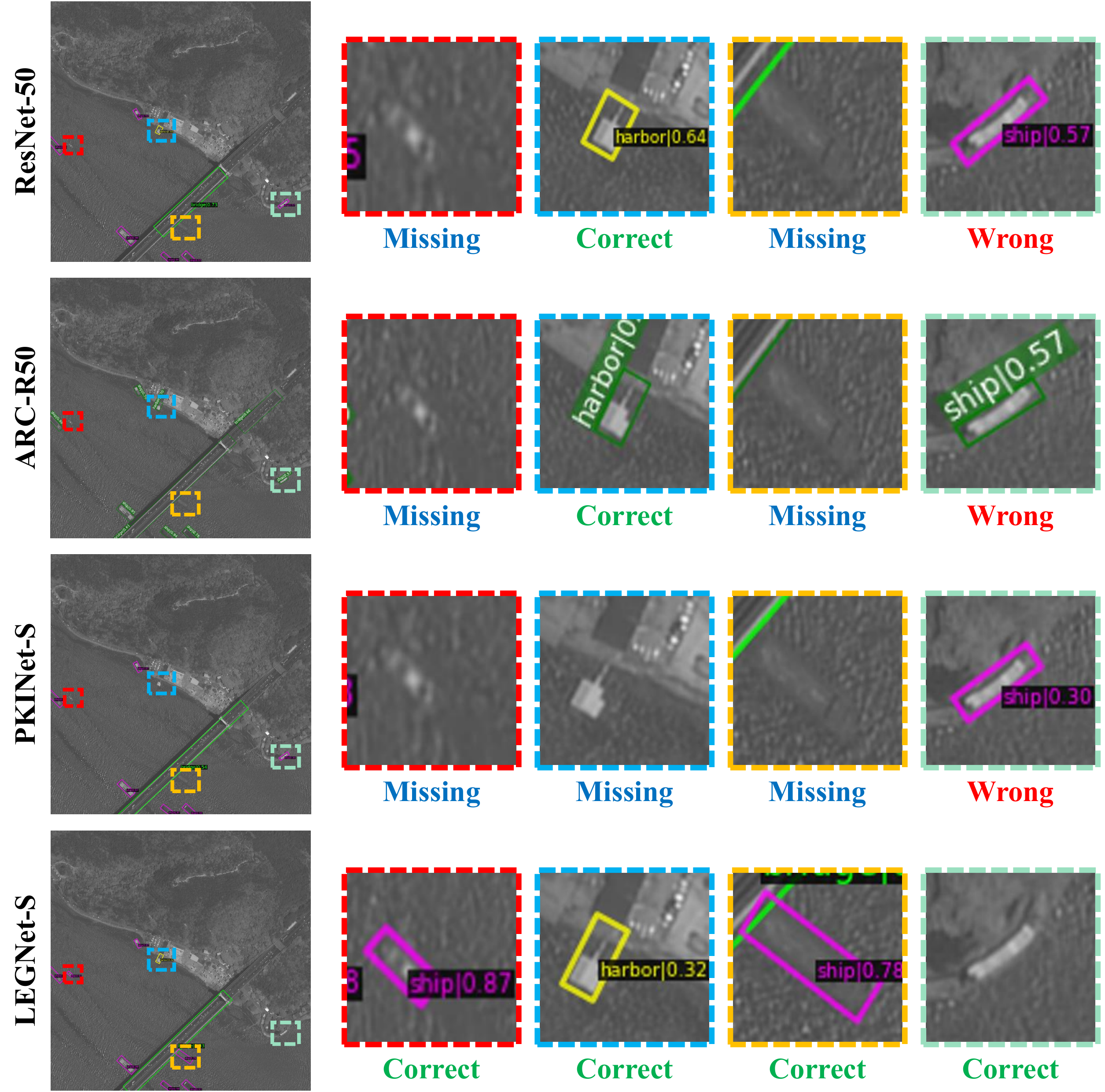}
	\end{center}	\vspace{-8pt}
	\caption{Visualization of detection results on DOTA-v1.0 test set~\cite{xia2018dota}. Input images resolution were 1,024 $\times$ 1,024.}	\vspace{-4pt}
	\label{comparison_visual}
\end{figure}

Visual results on the DOTA-v1.0 test set are presented in \cref{comparison_visual}. We visually compare LEGNet with SOTA backbones designed for RSOD: ARC-R50~\cite{pu2023adaptive} and PKINet~\cite{cai2024pkinet}. As depicted, LEGNet demonstrates superior capability in identifying blurry and low-quality objects, effectively detecting challenging examples such as ships. In contrast, ResNet-50, ARC-R50, and PKINet demonstrate suboptimal performance on such images. Furthermore, these methods exhibit false positives, for example, misclassifying a distinct white object on the shore as a ship, highlighting their limitations in precise object boundary discrimination.

\subsection{Ablation Studies} \label{ablation}

\begin{table}[t]\centering \scriptsize	
\setlength\tabcolsep{4pt}
\renewcommand{\arraystretch}{1.2}
\begin{tabular}{cc|ccc|cccc}	\specialrule{0.8pt}{0pt}{0pt}
\multirow{2}{*}{\makecell[c]{LoG- \\ -Stem}}  		
& \multirow{2}{*}{EGA} 
&\multirow{2}{*}{\#\textbf{P} $\downarrow$}
&\multirow{2}{*}{\textbf{FLOPs} $\downarrow$} 
&\multirow{2}{*}{\textbf{mAP}  $\uparrow$ } 
&\multicolumn{4}{c}{AP for each category ($\%$) $\uparrow$} \\  \cline{6-9}
&&&&&SV&LV&SP&HC \\ \hline 
&		 						&20.61 	&135.6	&70.8 &59.7&84.0&55.3&46.0\\
\ding{51} &		 				&20.65 	&148.5	&71.2 &61.5&84.8&61.3&52.7\\
&\ding{51} 						&20.61	&135.6	&71.5 &59.7&84.5&57.2&61.4\\ \hline
\rowcolor[rgb]{0.9,0.9,0.9} 
\multicolumn{2}{c}{LEGNet-T}\vline&20.65&148.5	&72.2 &63.2&85.0&59.6&58.3\\ \hline
\end{tabular} \vspace{-8pt}
\caption{Ablation study of LEGNet on DOTA-v1.0 validation set. Models were pre-trained 100 epoch on Imagenet-1k.} \label{Ablation}	\vspace{-5pt} 
\end{table}

\cref{Ablation} details an ablation study of LEGNet on the DOTA-v1.0 validation set. All models used a backbone pre-trained for 100 epochs on ImageNet-1K~\cite{deng2009imagenet}, following~\cite{li2023large,cai2024pkinet} for efficiency. We compare a baseline (no LoG-Stem or EGA), variants with only LoG-Stem or EGA, and the LEGNet-T. Ablating the EGA module means its fixed kernels for edge/Gaussian attention are replaced by learnable convolution. Ablating the LoG-Stem replaces it with a learnable \(4 \times 4\) stride-4 convolutional layer. We report mAP and AP for challenging categories: small vehicles (SV), large vehicles (LV), swimming pools (SP), and helicopters (HC).

As shown in \cref{Ablation}, the variant with only LoG-Stem (EGA removed) achieves 71.2\% mAP. The variant with only EGA (LoG-Stem removed) yields 71.5\% mAP, notably improving HC detection to 61.4\%. The full LEGNet-T attains the highest mAP of 72.2\%, significantly outperforming the baseline in SV (63.2\% vs. 59.7\%) and HC detection (58.3\% vs. 46.0\%). These results confirm that both LoG-Stem and EGA modules enhance detection performance, especially for these challenging classes.

\subsection{Limitations and Future Works}   \label{lim_future}
The optimal Gaussian kernel size requires careful selection, as its effectiveness is likely dataset-dependent, varying with object characteristics. Furthermore, while LEGNet surpasses current SOTA methods with fewer parameters and FLOPs, its inference speed indicates clear potential for further optimization (less optimized implementations compared to convolution in deep learning framework).

Future work could extend this framework to downstream tasks like semantic segmentation and object tracking. Leveraging its demonstrated robust feature extraction and efficiency, these applications could benefit from edge-Gaussian modeling. Extending our approach broadens its applicability and opens avenues for high performance in diverse vision challenges. Further exploration of task-specific adaptations and further optimization to build efficient and scalable systems for real-world problems.



%% file: sec/5con.tex
\section{Conclusion}
In this paper, we present LEGNet—a lightweight network designed for robust RSOD under low-quality imaging conditions. By bridging traditional image processing (orientation-aware Scharr edge enhancement and Gaussian-prior-based feature modeling) and deep learning, LEGNet enhances low-quality feature representation and object boundary clarity, achieving SOTA performance on multiple benchmarks while maintaining computational efficiency. Future works could extend this framework to additional downstream tasks, providing a versatile foundation for further research in aerial and satellite image analysis.
